\documentclass[journal]{IEEEtran}
\usepackage[usenames]{color}
\usepackage{amsmath,amssymb,graphicx,url, epsfig}
\usepackage{cite, subfigure, xspace, paralist}
\usepackage{bbm}
\usepackage{bm}

\def \tstc {\texttt{tst2013}\xspace}
\def \tsta {\texttt{tst2010}\xspace}
\def \tstb {\texttt{tst2011}\xspace}
\def \dev {\texttt{dev2010}\xspace}
\def \amidev {\texttt{dev}\xspace}
\def \amieval {\texttt{eval}\xspace}
\def \swbdeval {\texttt{eval2000}\xspace}
\def \lhuc {\texttt{LHUC}\xspace}
\def \LHUC {\texttt{LHUC}\xspace}

\def \fmllr {fMLLR\xspace}

\def \T {\top}

\def \f {\phi}
\def \nlf #1{\f^{#1}}

\def \diffg {\texttt{Diff-Gauss}\xspace} 
\def \diffp {\texttt{Diff-$L_p$}\xspace} 
\def \difff {\texttt{Diff-$L_{2}$}\xspace} 

\def \llpf {\diffp} 
\def \flpf {\difff} 
\def \gauss {\diffg}

\def \updateu { Update ${\mu}$}
\def \updateb { Update ${\beta}$}
\def \updatee { Update ${\eta}$}
\def \updateg { \footnotesize{Update $\mu, \beta$}}
\def \updatec { Update ${b}$}
\def \updatep { Update ${p}$}

\def \ANN {\mbox{DNN}\xspace}

\renewcommand{\vec}[1]{\mathbf{#1}}

\begin{document}

\title{Differentiable Pooling for Unsupervised \\ Acoustic Model Adaptation}

\author{Pawel~Swietojanski,~\IEEEmembership{Member,~IEEE,}\thanks{P Swietojanski and S Renals are with The Centre for Speech Technology Research, University of Edinburgh, Edinburgh EH89AB, U.K., (e-mail: p.swietojanski@ed.ac.uk, s.renals@ed.ac.uk)}
  and~Steve~Renals,~\IEEEmembership{Fellow,~IEEE}\thanks{The authors were supported by EPSRC Programme Grant grant EP/I031022/1 \emph{Natural Speech Technology} (NST) and the European Union under H2020 project \emph{SUMMA}, grant agreement 688139.}
}



%

\IEEEpubid{10.1109/TASLP.2016.2584700~\copyright~2016 IEEE.}
\markboth{IEEE/ACM Transactions on Audio, Speech and Language Processing, Vol. 24, Num. 11}{}


\maketitle

\begin{abstract}
We present a deep neural network (DNN) acoustic model that includes parametrised and differentiable pooling operators. Unsupervised acoustic model adaptation is cast as the problem of updating the decision boundaries implemented by each pooling operator. In particular, we experiment with two types of pooling parametrisations: learned $L_p$-norm pooling and weighted Gaussian pooling, in which the weights of both operators are treated as speaker-dependent. We perform investigations using three different large vocabulary speech recognition corpora: AMI meetings, TED talks and Switchboard conversational telephone speech. We demonstrate that differentiable pooling operators provide a robust and relatively low-dimensional way to adapt acoustic models, with relative word error rates reductions ranging from 5--20\% with respect to unadapted systems, which themselves are better than the baseline fully-connected DNN-based acoustic models. We also investigate how the proposed techniques work under various adaptation conditions including the quality of adaptation data and complementarity to other feature- and model-space adaptation methods, as well as providing an analysis of the characteristics of each of the proposed approaches.
\end{abstract}

%
\IEEEpeerreviewmaketitle

\section{Introduction and Summary}

\IEEEPARstart{D}{eep} neural network (DNN) acoustic models have significantly extended the state-of-the-art in speech recognition \cite{Hinton2012} and  are known to be able to learn significant invariances through many layers of non-linear transformations~\cite{Yu2013_repr}.  If the training and deployment conditions of the acoustic model are mismatched then the runtime data distribution can differ from the training distribution, bringing a degradation in accuracy, which may be addressed through explicit adaptation to the test conditions \cite{Neto1995, li2010lin, jan2010speaker, Seide2011, Yao2012, Yu2013_repr, Swietojanski2013, Yu2013}.  

In this paper we explore the use of parametrised and differentiable pooling operators for acoustic adaptation.  We introduce the approach of differentiable pooling using speaker-dependent pooling operators, specifically $L_p$-norm pooling and weighted Gaussian pooling (Section~\ref{sec:diffp}), showing how the pooling parameters may be optimised by minimising the negative log probability of the class given the input data (Section~\ref{sec:difflearn}), and providing a justification for the use of pooling operators in adaptation (Section~\ref{sec:whydiff}). To evaluate this novel adaptation approach we performed experiments on three corpora -- TED talks, Switchboard conversational telephone speech, and AMI meetings -- presenting results on using differentiable pooling for speaker independent acoustic modelling, followed by unsupervised speaker adaptation experiments  in which adaptation of the pooling operators is compared (and combined) with learning hidden unit contributions (LHUC) \cite{Abdel-Hamid2013_is, Swietojanski2014_lhuc} and constrained/feature-space maximum likelihood linear regression (fMLLR)~\cite{Gales1998}.

\IEEEpubidadjcol
\section{DNN Acoustic Modelling and Adaptation} 
\label{sec:review}

DNN acoustic models typically estimate the posterior distribution over a set  of context-dependent tied states $s$ of a hidden Markov model (HMM)~\cite{young1994tied} given an acoustic observation $\vec o$, $P(s | \vec o)=\ANN(\vec o; \bm \theta)$~\cite{Bourlard1994,Renals1994,Hinton2012}. The \ANN is implemented as a nested function comprising $L$ processing layers (non-linear transformations):
\begin{flalign} \label{eq:dnn}
  \ANN(\vec o; \bm \theta) = f^L \left (f^{L-1}\left (\ldots f^1 \left (\vec o; \theta^1 \right ) \ldots ;\theta^{L-1} \right ) ;\theta^L \right ) 
\end{flalign}
\noindent The model is thus parametrised by a set of weights $\bm \theta=\{\bm \theta^l\}_{l=1}^L$ in which the $l$th layer consists of a weight matrix and bias vector, $\bm \theta^l=\{\vec W^l, \vec b^l\}$, followed by a non-linear transformation $\nlf{}$, acting on arbitrary input $\vec x$:
\begin{equation}
f^l(\vec x; \bm \theta^l)=\nlf{l}\left (\vec W^{l\top}\vec x + \vec b^l \right )
\end{equation}
To form a probability distribution, the output layer employs a \emph{softmax} transformation~\cite{Bridle1990softmax} $\nlf{L}_i(\vec x)=\exp(x_i)/\sum_{j}\exp(x_{j})$,
 whereas the hidden layer activation functions are typically chosen to be either sigmoid $\nlf{l}(x)=1/(1+\exp(-x))$  or rectified linear $\nlf{l}(x)=\max(0, x)$ units (ReLU)~\cite{Nair2010}.

Yu et al~\cite{Yu2013_repr} experimentally demonstrated that  
the invariance of the internal representations with respect to variabilities in the input space increases with depth (the number of layers) and that the DNN can interpolate well around training samples but fails to extrapolate if the data mismatch increases. Therefore one often explicitly compensates for unseen variabilities in the acoustic space. 

\emph{Feature-space normalisation} increases the invariance to unseen data by transforming the data such that it better matches the training data. In this approach the DNN learns an additional  transform of the input features conditioned on the speaker or the environment.  The transform, which is typically affine, is parametrised by an additional set of adaptation parameters.  The most effective form of feature-space normalisation is constrained (feature-space) maximum-likelihood linear regression (MLLR), referred to as fMLLR~\cite{Gales1998}, in which the linear transform parameters are estimated by maximising the likelihood of the adaptation data under a Gaussian Mixture Model (GMM) / HMM acoustic model.  To use fMLLR with a DNN acoustic model it is necessary to estimate a single input transform per speaker (using a trained GMM),  using the resultant transformed data to train a DNN in a speaker adaptive training (SAT) manner.  At runtime another set of fMLLR parameters is estimated for each speaker and the data transformed accordingly.  This technique has consistently and significantly reduced the word error rate (WER) across several different benchmarks for both hybrid~\cite{Bourlard1994, Hinton2012} and tandem\cite{Hermansky2000, Grezl2007} approaches.  There are many successful examples of fMLLR adaptation of DNN acoustic models~\cite{Mohamed2011, Seide2011, Hain2012, Hinton2012, Sainath2012, Sainath2013_cnns, Bell2013_mlan, Swietojanski2013, yoshioka2014investigation}.  One can also estimate the linear transform as an input layer of the DNN, often referred to as a linear input network (LIN)~\cite{Neto1995, Abrash1995,li2010lin, Seide2011}. LIN-based approaches have been mostly used in test-only adaptation schemes, whereas fMLLR requires SAT, but usually results in lower WERs.

\emph{Auxiliary feature} approaches augment the acoustic feature vectors with additional speaker-dependent information computed for each speaker at both training and runtime stages -- this is a form of SAT in which the model learns the distribution over tied states conditioned on some additional speaker-specific information.  There has been considerable recent work exploring the use of i-vectors~\cite{Dehak2010} for this purpose.  I-vectors, which can be regarded as  basis vectors spanning a subspace of speaker variability, were first used for adaptation in a GMM framework by Karafiat et al~\cite{Karafiat2011}, and were later successfully employed for DNN adaptation ~\cite{Saon2013, senior2014adapt, Gupta2014, Karanasou2014, Miao2015,  Samarakoon:ICASSP16}.
Other examples of auxiliary features include the use of speaker-specific bottleneck features obtained from a speaker separation DNN~\cite{Liu2014}, the use of out-of-domain tandem features~\cite{Bell2013_mlan}, as well as speaker codes~\cite{Bridle1990, Abdel-Hamid2013, Xue2014_scodes} in which a specific set of units for each speaker is optimised. Kundu et al.~\cite{kundu2016factor} present an approach using auxiliary input features derived from the bottleneck layer of a DNN which is combined with i-vectors.

\emph{Model-based} approaches adapt the DNN parameters using data from the target speaker.   Liao~\cite{Liao2013} investigated this approach in both supervised and unsupervised settings using a few minutes of adaptation data.  On a large DNN, when all weights were adapted, up to 5\% relative improvement was observed for unsupervised adaptation, using a speaker independent decoding to obtain DNN targets.  Yu et al~\cite{Yu2013} have explored the use of regularisation for adapting the weights of a DNN,  using the Kullback-Liebler (KL) divergence between the output distributions produced by speaker-independent and the speaker-adapted models.  This approach was also recently used to adapt parameters of sequence-trained models~\cite{Huang2015}.  
LIN may also be regarded as a form of model-based adaptation, and related approaches include adaptation using  a linear output network (LON) or linear hidden network (LHN)~\cite{li2010lin, Yao2012, Ochiai2014}.

Directly adapting all the weights of a large DNN is computationally and data intensive, and results in large speaker-dependent parameter sets.  Smaller subsets of the DNN weights may be modified, including biases and slopes of hidden units~\cite{Sini2013, Yao2012, Zhao2015_slopes, Samarakoon:ICASSP16}.
Another recently developed approach relies on learning hidden unit contributions (LHUC) for test-only adaptation~\cite{Abdel-Hamid2013_is, Swietojanski2014_lhuc} as well as in a SAT framework~\cite{Swietojanski:ICASSP16}.  One can also adapt the top layer using Bayesian methods resulting in a maximum a posteriori (MAP) approach~\cite{huang2015maximum}, or address the sparsity of context-dependent tied-states when few adaptation data-points are available by modelling both monophones and context-dependent tied states using multi-task adaptation~\cite{Huang2015_mt, Swietojanski2015_mt} or a hierarchical output layer~\cite{Price2014}.

\section{Differentiable Pooling} \label{sec:diffp}

\def \Rk {\ensuremath{R_k}}
\def \hk {g}
\def \ak {\vec a^k} 
\def \aki {a^k_i} 
\def \ake #1{a^k_{#1}} 
\def \pk {p_k}
\def \rpsum { \sum_{i \in \Rk} \vert \aki \vert^{\pk}} 
\def \rpnorm { \frac{1}{K}}

\def \zk {\vec z^{k}} 
\def \zkt {(\zk)^\top}
\def \zki {z^k_i}
\def \zkip {z^k_{i'}}
\def \gp {\theta^k}

\def \jacu {\vec J_{\vec u}(\vec v(\zk))}
\def \jacv {\vec J_{\vec v}(\zk)}

\def \dudv[#1]#2{\frac{\partial u(z^k_{#1})}{\partial v(z^k_{#2})}}
\def \dvdz[#1]#2{\frac{\partial v(z^k_{#1})}{\partial z^k_{#2}}}

\def \jacm {\vec J_{\vec v}(\mu_k)}
\def \jacb {\vec J_{\vec v}(\beta_k)}

Building on our initial work~\cite{Swietojanski:ICASSP15}, we present an approach to adaptation by learning hidden layer pooling operators with parameters that can be learned and adapted in a similar way to the other model parameters.  The idea of feature pooling originates from Hubel and Wiesel's pioneering study on visual cortex in cats~\cite{hubel1962receptive}, and was first used in computer vision to combine spatially local features~\cite{Fukushima1982}.  Pooling in DNNs involves the combination of a set of hidden unit outputs into a summary statistic.  Fixed poolings are typically used, such as average pooling (used in the original formulation of convolutional neural networks -- CNNs)~\cite{LeCun1989,LeCun1998a} and max pooling (used in the context of feature hierarchies~\cite{Riesenhuber1999} and later applied to CNNs~\cite{Ranzato2007,Boureau2010}).  

Reducing the dimensionality of hidden layers by pooling some subsets of hidden unit activations has become well investigated beyond computer vision, and the $\max$ operator has been interpreted as a way to learn piecewise linear activation functions -- referred to as Maxout~\cite{Goodfellow2013}.  Maxout has been widely investigated for both fully-connected~\cite{Miao2013b, Cai2013, swietojanski2014maxout} and convolutional~\cite{Renals2014, Toth2014} DNN-based acoustic models.  Max pooling, although differentiable, performs a one-from-$K$ selection, and hence does not allow hidden unit outputs to be interpolated, or their combination to be learned within a pool. 

There have been a number of approaches to pooling with differentiable operators -- \emph{differentiable pooling} -- a notion introduced by Zeiler and Fergus~\cite{Zeiler2012} in the context of constructing unsupervised feature extract for support vector machines in computer vision tasks.  There has been some interest in the use of $L_p$-norm pooling with CNN models~\cite{Boureau2010, Sermanet2012} in which the sufficient statistic is the $p$-norm of the group of (spatially-related) hidden unit activations. Fixed order $L_p$-norm pooling was recently applied within the context of a convolutional neural network acoustic model~\cite{Sainath2013}, where it did not reduce the WER over max-pooling, and as an activation function in a fully-connected DNNs~\cite{Zhang2014}, where it was found to improve over maxout and ReLU. 

\begin{figure}
\centering
\includegraphics[width=0.9\columnwidth]{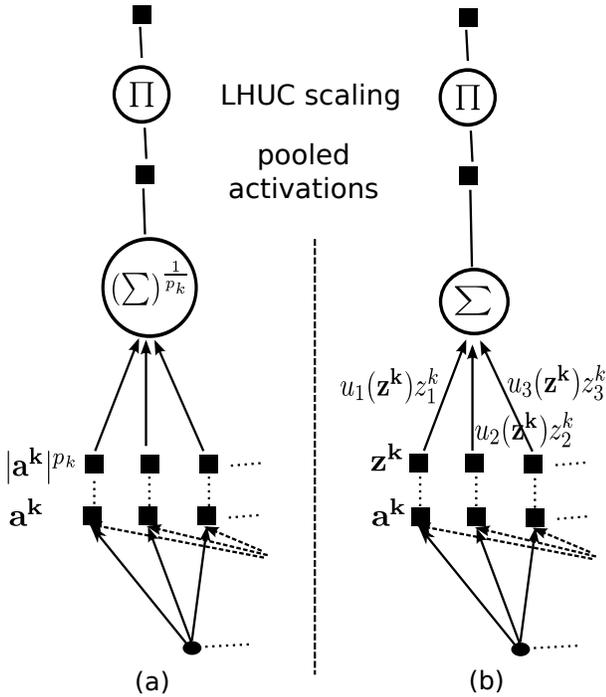}
\caption{Illustration of \diffp (a) and \diffg (b) pooling operators. See Sections~\ref{sec:lp} and~\ref{sec:gauss} for further details and explanations of the symbols. \lhuc scaling follows~\cite{swietojanski2016lhuc} and is used only during adaptation.}
\label{fig:diffp_nnets}
\end{figure}

\subsection{$L_p$-norm (\diffp) pooling} \label{sec:lp}

\def \hk {f_{L_p}}
\def \ak {\vec a^k} 
\def \aki {a^k_i} 
\def \ake #1{a^k_{#1}} 
\def \pk {p_k}
\def \prep {\zeta}
\def \rpsum { \sum_{i \in \Rk} \vert \aki \vert^{\pk}} 
\def \rpnorm { \frac{1}{K}}

In this approach we pool a set of activations using an $L_p$-norm.  A hidden unit pool is formed by a set $\Rk$ of $K$ affine projections which form the input to the $k$th pooling unit, which we write as an ordered set (vector) $\ak = \{ \vec w^{\top}_i \vec x + b_i \}_{i\in \Rk}$.
The  output of the $k$th pooling unit is produced as an $L_p$ norm:
\begin{equation}
\hk\left ( \ak; \pk \right ) = ||\ak||_{\pk} = \left ( \rpnorm \rpsum \right )^{\frac{1}{\pk}},
\label{eq:lp}
\end{equation}
\noindent where $\pk$ is the learnable norm order for the $k$th unit, that can be jointly optimised with the other parameters in the model. To ensure that~\eqref{eq:lp} satisfies a triangle inequality ($p_k\geq 1$; a necessary property of the norm), during optimisation $p_k$ is re-parametrised as $\pk = \prep(\rho_k) = \max(1,\rho_k)$, where $\rho_k$ is the actual learned parameter. For the case when $\pk=\infty$ we obtain the max-pooling operator \cite{Riesenhuber1999}:
\begin{equation}
||\ak||_{\infty} = \max \left ( \{|\ake{i}|\}_{i\in \Rk} \right ).
\label{eq:lpinf}
\end{equation}
Similarly, if $\pk=1$ we obtain absolute average pooling (assuming the pool is normalised by $K$). We refer to this model as \diffp, and it is parametrised by $\theta_{L_p}=\{\{\vec W^l, \vec b^l, \pmb \rho^l\}_{l=1}^{L-1}, \vec W^L, \vec b^L\}$. Serement et al~\cite{Sermanet2012} investigated fixed-order $L_p$ pooling for image classification, which was applied to speaker independent acoustic modelling~\cite{Zhang2014}.  Here we allow each $L_p$ unit in the model to have a learnable order $p$~\cite{Caglar2013}, and we use the pooling parameters to perform model-based test-only acoustic adaptation. 

\subsection{Gaussian kernel (\diffg) pooling} \label{sec:gauss}

\def \gk {f_{G}}
\def \zk {\vec z^{k}} 
\def \zkt {(\zk)^\top}
\def \zki {z^k_i}
\def \zkip {z^k_{i'}}
\def \gp {\bm \vartheta^k}

\def \jacu {\vec J_{\vec u}(\vec v(\zk))}
\def \jacv {\vec J_{\vec v}(\zk)}

\def \dudv[#1]#2{\frac{\partial u(z^k_{#1})}{\partial v(z^k_{#2})}}
\def \dvdz[#1]#2{\frac{\partial v(z^k_{#1})}{\partial z^k_{#2}}}

\def \jacm {\vec J_{\vec v}(\mu_k)}
\def \jacb {\vec J_{\vec v}(\beta_k)}
The second pooling approach estimates the pooling coefficients using a Gaussian kernel.  We generate the pooling inputs at each layer as:  
\begin{equation}
\zk = \left \{ \eta_k \cdot \nlf{}(\vec w^{\top}_i \vec x + b_i) \right \}_{i\in \Rk} = \left \{ \eta_k \cdot \nlf{}(\vec a^{k}_i) \right \}_{i\in \Rk},
\label{eq:nonlin}
\end{equation}
\noindent where $\nlf{}$ is a non-linearity ($\tanh$ in this work) and $\ak$ is a set of affine projections as before. A non-linearity is essential as otherwise (contrary to $L_p$ pooling) we would produce a linear transformation through a linear combination of linear projections. $\eta_k$ is the pool amplitude; this parameter is tied and learned per-pool as this was found to give similar results to per-unit amplitudes (but with fewer parameters), and better results compared to setting to a fixed value $\eta_k=1.0$~\cite{Swietojanski:ICASSP15}.

Given the activation \eqref{eq:nonlin}, the pooling operation is defined as a weighted average over a set $\Rk$ of hidden units, where the $k$-th pooling unit $\gk(\cdot; \gp)$ is expressed as:
\begin{equation} \label{eq:gp_pool}
 \gk \left ( \zk ; \gp \right ) = \sum_{i \in \Rk} u_i(\zk; \gp)\zki.
 \vspace{-2mm}
\end{equation}
\noindent The pooling contributions $\vec u(\zk; \gp)$ are normalised to sum to one within each pooling region $\Rk$ \eqref{eq:weight}  and each weight $u_i(\zki; \gp)$ is coupled with the corresponding value of $\zki$ by a Gaussian kernel \eqref{eq:pdf} (one per pooling unit) parameterised by the mean and precision, $\gp=\{\mu_k, \beta_k\}$: 
\begin{equation} \label{eq:weight}
 u_i(\zk; \gp) = \frac{v(\zki; \gp)}{\sum_{i' \in \Rk} v(\zkip; \gp)},
\end{equation}
\begin{equation} \label{eq:pdf}
 v(\zki; \gp) = \exp \left (-\frac{\beta_k}{2}\left (\zki-\mu_k \right )^2 \right ).
\end{equation}
Similar to $L_p$-norm pooling, this formulation allows a generalised pooling to be learned -- from average ($\beta \rightarrow 0$) to $\max$ ($\beta \rightarrow \infty$) -- separately for each pooling unit $\gk(\zk;\gp)$ within a model. The \diffg model is thus parametrised by $\theta_{G}=\{\{\vec W^l, \vec b^l, \pmb \mu^l, \pmb \beta^l, \pmb \eta^l\}_{l=1}^{L-1}, \vec W^L, \vec b^L\}$.

\section{Learning Differentiable Poolers}\label{sec:difflearn}

We optimise the acoustic model parameters by minimising the negative log probability of the target HMM tied state given the acoustic observations using gradient descent and error back-propagation~\cite{rumelhart1986learning}; the pooling parameters may be updated in a speaker-dependent manner, to adapt the acoustic model to unseen data. In this section we give the necessary partial derivatives for \diffp and \diffg pooling.

\subsection{Learning and adapting \diffp pooling}
In \diffp pooling we learn $p_k$ which we express in terms of $\rho$, $\pk=\prep(\rho_k)$.  Error back-propagation requires the partial derivative of the pooling region $\hk(\ak; \rho_k)$ with respect to $\rho_k$, which is given as:
\begin{flalign} \label{eq:lp_p}
\frac{\partial \hk(\ak; \rho_k)}{\partial \rho_k} = \Bigg ( & \frac{\sum_{i\in \Rk} \log(|\aki|)\cdot|\aki|^{\pk} }{\pk\rpsum} \\ 
& - \frac{\log\rpsum}{\pk^2} \Bigg )\frac{\partial \prep(\rho_k)}{\partial \rho_k}\hk(\ak; \rho_k), \nonumber
\end{flalign}
\noindent where $\partial \prep(\rho_k) / \partial \rho_k = 1$ when $\pk > 1$ and 0 otherwise. The back-propagation through the norm itself is implemented as:
\begin{align}
\frac{\partial \hk (\ak; \pk)}{\partial \ak} =  \frac{\ak \circ |\ak|^{\pk-2}}{ \rpsum } \circ \vec G^k,
\label{eq:lp_a}
\end{align}
\noindent where $\circ$ represents the element-wise Hadamard product, and $\vec G^k$ is a vector of $\hk(\ak; \pk)$ activations repeated $K$ times, so the resulting operation can be fully vectorised:
\begin{align} \label{eq:lp_g}
	\vec G^k = \left [ \hk( \ak; \pk)^{1}, \ldots, \hk( \ak; \pk)^{K} \right ]^\top.
\end{align}
Normalisation by $K$ in~\eqref{eq:lp} is optional (see also Section~\ref{ssec:base}) and the partial derivatives in \eqref{eq:lp_p} and~\eqref{eq:lp_a} hold for the un-normalised case also: the effect of this is taken into account in the forward activation $\hk(\ak; \pk)$. 

Since \eqref{eq:lp_p} and \eqref{eq:lp_a} are not continuous everywhere, they need to be stabilised when $\rpsum=0$. When computing logarithm in the numerator of~\eqref{eq:lp_p} it is also necessary to ensure that each $\aki>0$. In practise, we threshold each element to have at least a value $\epsilon=10^{-8}$ if $\aki<\epsilon$. Note, this numerical stabilisation of $\ak$ only applies to $L_p$ units, not \diffg.

\subsection{Learning and adapting \diffg pooling regions}
To learn the \diffg pooling parameters $\gp=\{\mu^k,\beta^k\}$, we require the partial derivatives $\partial \gk(\zk)/ \partial \mu_k$ and $\partial \gk(\zk)/ \partial \beta_k$ to update pooling parameters, as well as $\partial \gk(\zk)/ \partial \zk$ in order to back-propagate error signals to lower layers. 

One can compute the partial derivative of \eqref{eq:gp_pool} with respect to the input activations $\zk$ as:
\begin{flalign} \label{eq:dz}
  \frac{\partial \gk(\zk)}{\partial \zk} = \left[\zkt \left ( \jacu\jacv \right ) + \vec u(\zk)^\T\right]^\top,
\end{flalign}
\noindent where $\jacu$ is the Jacobian representing the partial derivative $\partial u(\zk)/\partial v(\zk)$:
\begin{equation} \label{eq:jacu}
\jacu = \frac{\partial \vec u(\zk)}{\partial \vec v(\zk)} =
\begin{bmatrix} \dudv[1]{1} & \cdots & \dudv[1]{K} \\ \vdots & \ddots & \vdots \\ \dudv[K]{1} & \cdots & \dudv[K]{K} \end{bmatrix},
\end{equation}
\noindent whose elements can be computed as:
\begin{equation} \label{eq:dw1}
 \frac{\partial u(\zki)}{\partial v(\zki)} = \Big (\sum_{m\in \Rk} v(z_m) \Big )^{-1}\left (1 - u(\zki) \right),
\end{equation}
\begin{equation} \label{eq:dw2}
 \frac{\partial u(\zki)}{\partial v(\zkip)} = \Big (\sum_{m\in \Rk} v(z_m) \Big )^{-1}\left (-u(\zki) \right).
\end{equation}
\noindent Likewise, $\jacv$ represents the Jacobian of the kernel function $v(\zk)$ in \eqref{eq:pdf} with respect to $\zk$:
\begin{equation} \label{eq:jacv}
\jacv = \frac{\partial \vec v(\zk)}{\partial \zk} =
\begin{bmatrix} \dvdz[1]{1} & \cdots & 0 \\ \vdots & \ddots & \vdots \\ 0 & \cdots & \dvdz[K]{K} \end{bmatrix},
\end{equation}
\noindent and the elements of $\jacv$ can be computed as:
\begin{equation} \label{eq:dvz}
 \frac{\partial v(\zki)}{\partial \zki} = -\beta_k(\zki-\mu_k)v(\zki).
\end{equation}
Similarly, one can obtain the gradients with respect to the pooling parameters $\gp$. In particular, for $\beta_k$, the gradient is:
\begin{flalign} \label{eq:dgdb}
  \frac{\partial \gk(\zk)}{\partial \beta_k} = \sum_{i\in \Rk} \left [ \zkt \left ( \jacu\jacb \right ) \right ]_i,
\end{flalign}
\noindent where $\jacb$ = $\partial \vec v(\zk)/\partial \beta_k$ and $\partial v(\zki)/\partial \beta_k$ is:
\begin{equation} \label{eq:dprec}
 \frac{\partial v(\zki)}{\partial \beta_k} = -\frac{1}{2}\left(\zki-\mu_k\right)^2v(z_i).
\end{equation}
The corresponding gradient for $\partial \gk(\zk)/ \partial \mu_k$ is obtained  below \eqref{eq:dgdm}. Notice, that $\partial v(\zki)/\partial \zki$ \eqref{eq:dvz} and $\partial v(\zki)/\partial \mu_k$ \eqref{eq:dmean} are symmetric, hence $\jacm = -\jacv$, and to compute $\partial \gk(\zk)/ \partial \mu_k$ one can reuse the $\zkt\jacu\jacv$ term in~\eqref{eq:dz}, as follows:
\begin{flalign} \label{eq:dgdm}
  \frac{\partial \gk(\zk)}{\partial \mu_k} &= \sum_{i\in \Rk} \left [ \zkt \left ( \jacu\jacm \right ) \right ]_i \nonumber \\
  &= - \sum_{i\in \Rk} \left [\zkt \left ( \jacu\jacv \right ) \right ]_i,
\end{flalign}
\begin{equation} \label{eq:dmean}
 \frac{\partial v(\zki)}{\partial \mu_k} =  -\frac{\partial v(\zki)}{\partial \zki} = \beta_k(\zki - \mu_k)v(\zki).
\end{equation}
\section{Representational efficiency of pooling units} \label{sec:whydiff}

\begin{figure}
\centering
\includegraphics[width=1.0\columnwidth]{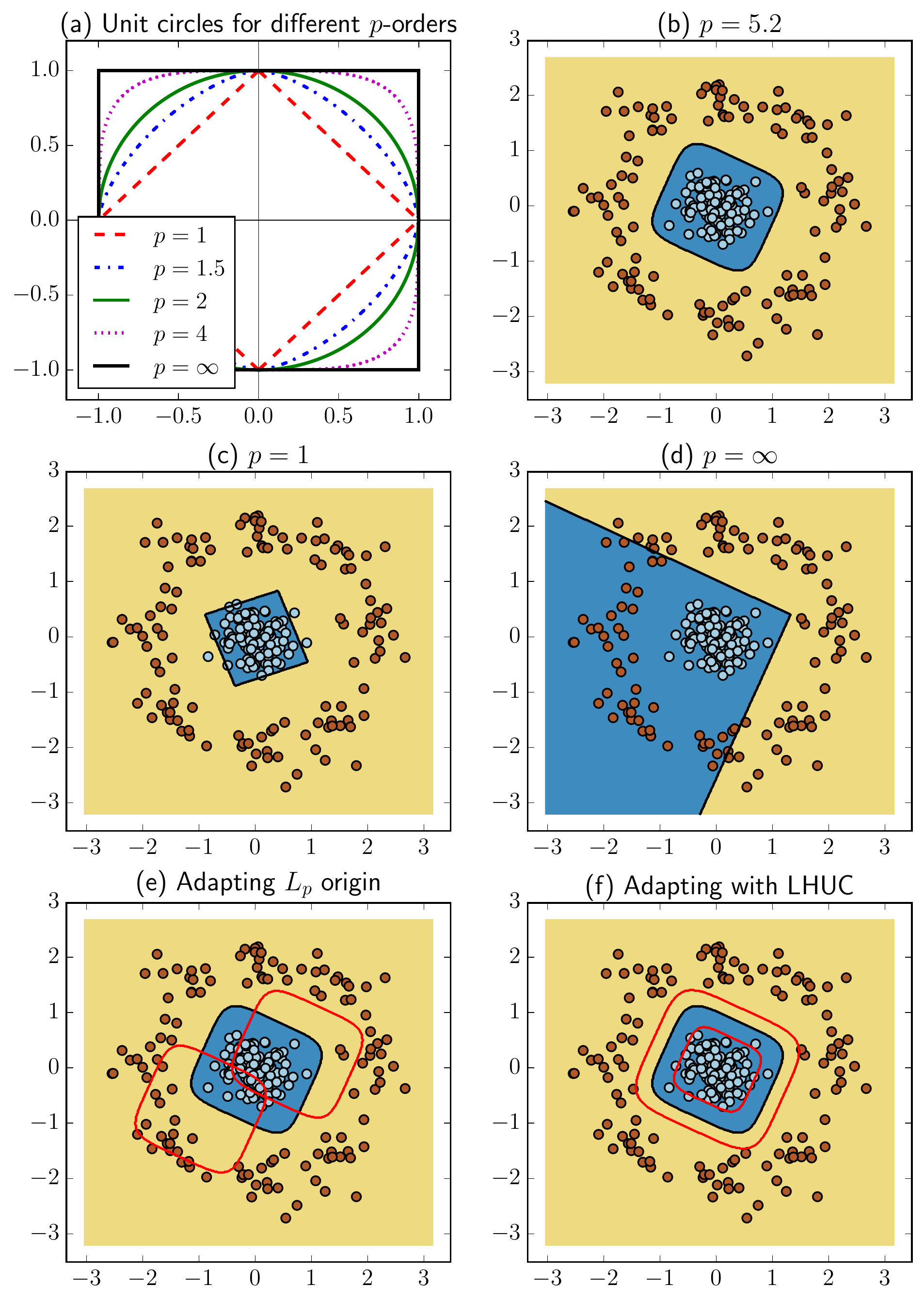}
\caption{Illustration of the representational efficiency and adaptation principles of an $L_p$ unit. (a) Unit circles as obtained under different norm $p$-orders. (b) An example decision boundary for two-class toy data (red and blue dots). The \diffp model is build out of one $L_p$ unit (with $K=2$ linear inputs) and is able to draw highly non-linear decision-regions. (c) The model from (b) with $p = 1.0$ and (d) $p=\infty$. Red contours of the bottom two plots illustrate (e) the effect of adaptation of the origin (biases) of the linear inputs $\ak$ and (f) the effect of \lhuc scaling. Further description in Section~\ref{sec:whydiff}. (Best viewed in colour.)}
\label{fig:diffp_regions}
\end{figure}

The aim of model-based DNN adaptation is to alter the learned speaker independent representation in order to improve the classification accuracy for data from a possibly mismatched test distribution. Owing to the highly distributed representations that are characteristic of DNNs, it is rarely clear which parameters should be adapted in order generalise well to a new speaker or acoustic condition.

Pooling  enables decision boundaries to be altered, through the selection of relevant hidden features, while keeping the parameters of the feature extractors (the hidden units) fixed: this is similar to \lhuc adaptation~\cite{swietojanski2016lhuc}.  The pooling operators allow for a geometrical interpretation of the decision boundaries and how they will be affected by a constrained adaptation -- the units within the pool are jointly optimised given the pooling parametrisation, and share some underlying relationship within the pool. 

This is visualised  for $L_p$ units in Fig.\ \ref{fig:diffp_regions}. Fig.\ \ref{fig:diffp_regions} (a) illustrates the unit circles obtained by solving $||\ak||_{p}=d$ for different orders $p$, with $d=1.0$ and a pool of $K=2$ linear inputs $\ak$.  Such an $L_p$ unit is capable of closed-region decision boundaries, illustrated in Fig.\ \ref{fig:diffp_regions} (b).  
The distance threshold $d$ is implicitly learned from data (through the $\ak$ parameters given $p$),  
resulting in an efficient representation~\cite{Caglar2013, Zhang2014} compared with representing such boundaries using sigmoid units or ReLUs, which would require more parameters. Figs.\ \ref{fig:diffp_regions} (c) and (d) show how those boundaries are affected when $p=1$ (average pooling) and $p=\infty$ (max pooling), while keeping $\ak$ fixed. As shown in Section~\ref{sec:results} we found that updating $p$ is an efficient and relatively low-dimensional way to adjust decision boundaries such that the the model's accuracy on the adaptation data distribution improves.  

It is also possible to update the biases (Fig.\ \ref{fig:diffp_regions} (e), red contours) and the \lhuc amplitudes (Fig.\ \ref{fig:diffp_regions} (f), red contours).  We experimentally investigate how each approach impacts adaptation WER in Section~\ref{ssec:adapt}.  Although models implementing \diffg units are theoretically less efficient in terms of SI representations compared to $L_p$ units, and comparable to standard fully-connected models, the pooling mechanism still allows for more efficient (in terms of number of SD parameters) speaker adaptation.

\section{Experimental setups}  \label{sec:setups}
We have carried out experiments on three corpora:  the TED talks corpus~\cite{Cettolo2012} following the IWSLT evaluation protocol (\url{www.iwslt.org}); the Switchboard corpus of conversational telephone speech~\cite{godfrey1992switchboard} (\url{ldc.upenn.edu}) and the AMI meetings corpus \cite{Carletta_LRE2007, Renals_ASRU2007} (\url{corpus.amiproject.org}).  Unless explicitly stated otherwise, our baseline models share similar structure across the tasks -- DNNs with 6 hidden layers (2,048 units per layer) using a sigmoid non-linearity. The output softmax layer models the distribution of context-dependent clustered tied states~\cite{Dahl2012}. The features are presented in 11 ($\pm5$) frame long context windows. All the adaptation experiments, if not stated otherwise, were performed unsupervised using adaptation targets obtained from first-pass speaker-independent decoding of the corresponding SI system.

\textbf{TED}: 
The training data consisted of 143 hours of speech (813 talks) and the systems follow our previously described recipe~\cite{Swietojanski2013}. However, compared to our previous work~\cite{Swietojanski2013, Swietojanski2014_lhuc, Swietojanski:ICASSP15}, our systems here make use of more accurate language models developed for our IWSLT--2014 systems~\cite{Bell2014}: in particular, the final reported results use a 4-gram language model estimated from 751~million words. The baseline TED acoustic models were trained on unadapted PLP features with first and second order time derivatives.  We present results on four IWSLT test sets:  \dev, \tsta, \tstb and \tstc containing 8, 11, 8, and 28 talks respectively. 

\textbf{AMI}: We follow a Kaldi GMM recipe~\cite{Swietojanski_ASRU13} and use the individual headset microphone (IHM) recordings.  On this corpus, we train the acoustic models using 40 mel-filter-bank (FBANK) features. We decode with a pruned 3-gram language model estimated from 800k words of AMI training transcripts interpolated with an LM trained on Fisher conversational telephone speech transcripts (1M words) \cite{cieri2004}.

\textbf{Switchboard} (SWBD):  We follow a Kaldi GMM recipe~\cite{Vesely:IS13, Kaldi:ASRU11}\footnote{To stay compatible with our previous adaptation work on Switchboard~\cite{Swietojanski:ICASSP16, swietojanski2016lhuc} we are using the older set of Kaldi recipe scripts called \texttt{s5b}, and our baseline results are comparable with the corresponding baseline numbers previously reported. A newer set of improved scripts exists under \texttt{s5c} which, in comparison to \texttt{s5b}, offer about 1.5\% absolute lower WER.}, using Switchboard--1 Release 2 (LDC97S62). Our baseline unadapted acoustic models were trained on MFCC features, while the SAT trained fMLLR variants utilise the usual Kaldi feature preprocessing pipeline, which is MFCC+LDA/MLLT+fMLLR\footnote{MFCC-Mel-frequency Cepstral Coefficients, LDA - Linear Discriminant Analysis, MLLT - Maximum Likelihood Linear Transform}. The results are reported on the full Hub5'00 set (LDC2002S09) -- \swbdeval.  \swbdeval contains two types of data:  Switchboard -- which is better matched to the training data; and CallHome (CHE) English. Our reported results use 3-gram LMs estimated from the Switchboard and Fisher Corpus transcripts.
\section{Results} \label{sec:results}

\subsection{Baseline speaker independent models} \label{ssec:base}

The structures of the differentiable pooling models were selected such that the number of parameters was comparable to the corresponding baseline DNN models, described in detail in~\cite{swietojanski2016lhuc}. For the \diffp and \difff types, the resulting models utilised non-overlapping pooling regions of size $K=5$, with 900 $L_p$-norm units per layer.  The \diffg models had pool sizes set to $K=3$ (this was found to work best in our previous work~\cite{Swietojanski:ICASSP15}) which (assuming non-overlapping regions) results in 1175 pooling units per layer. 

\textbf{Training speaker independent \difff and \diffp models}: For both \diffp and \difff we trained with an initial learning rate of .008 (for MFCC, PLP, FBANK features) and .006 (for fMLLR features).  The learning rate was adjusted using the \texttt{newbob} learning scheme \cite{Renals1992} based on the validation  frame error rate.  We found that applying explicit pool normalisation (dividing by $K$ in~\eqref{eq:lp}) gives consistently higher error rates (typically an absolute increase of 0.3\% WER): hence we used un-normalised $L_p$ units in all experiments.  We did not apply post-layer normalisation~\cite{Zhang2014}.  Instead, we use max-norm approach -- after each update we scaled the columns (i.e. each $\aki$) of the fully connected weight matrices such that their $L_2$ norms were below a given threshold (set to $1.0$ in this work)~\cite{srivastava2014dropout}.  For \diffp models we initialised $p=2.0$. Those parameters were optimised on TED and directly applied without further tuning for the other two corpora.  In this work we have focussed on adaptation;  Zhang et al~\cite{Zhang2014} have reported further speaker independent experiments for fixed order $L_p$ units.

\textbf{Training speaker independent \diffg models}: The initial learning rate was set to 0.08 (regardless of the feature type), again adjusted using \texttt{newbob}. 
Initial pooling parameters were sampled randomly from normal distribution: $\mu\sim\mathcal{N}(0, 1)$ and  $\beta\sim\mathcal{N}(1, 0.5)$. Otherwise, the hyper-parameters were the same as for the baseline DNN models.

\textbf{Baseline speaker independent results}: Table \ref{tab:baselines} gives speaker independent results for each of the considered model types. The \diffg and \difff/\diffp models have comparable WERs, with a small preference towards \diffp in terms of the final WER on TED and AMI; all have lower average WER than the baseline DNN. The gap between the pooled models increases on AMI data where \diffp has a substantially lower WER (3.2\% relative) than the fixed order \difff which is in turn has a lower WER than the other two models (\diffg and baseline DNN) by 2.1\% relative.

\begin{table}[t]
\small
\caption{Baseline WER(\%) results on selected test sets of our benchmark corpora}
\label{tab:baselines}
\centerline{
\begin{tabular}{l||c|c|c}
 		& TED & AMI  & SWBD  \\ 
Model   & \tsta & \amieval & \swbdeval \\ 
\hline \hline
DNN			 & 15.0  &  29.1  & 22.1\\ 
\gauss 		 & 14.6	 &   29.0 & 21.4 \\ 
\difff		 &  14.6 &  28.5  & 21.3\\ 
\diffp		 &  14.5 &  27.6  & 21.3 \\ 
 \hline
\end{tabular}}
\end{table}

\begin{figure}[t]
 \centering  \includegraphics[width=1.0\columnwidth]{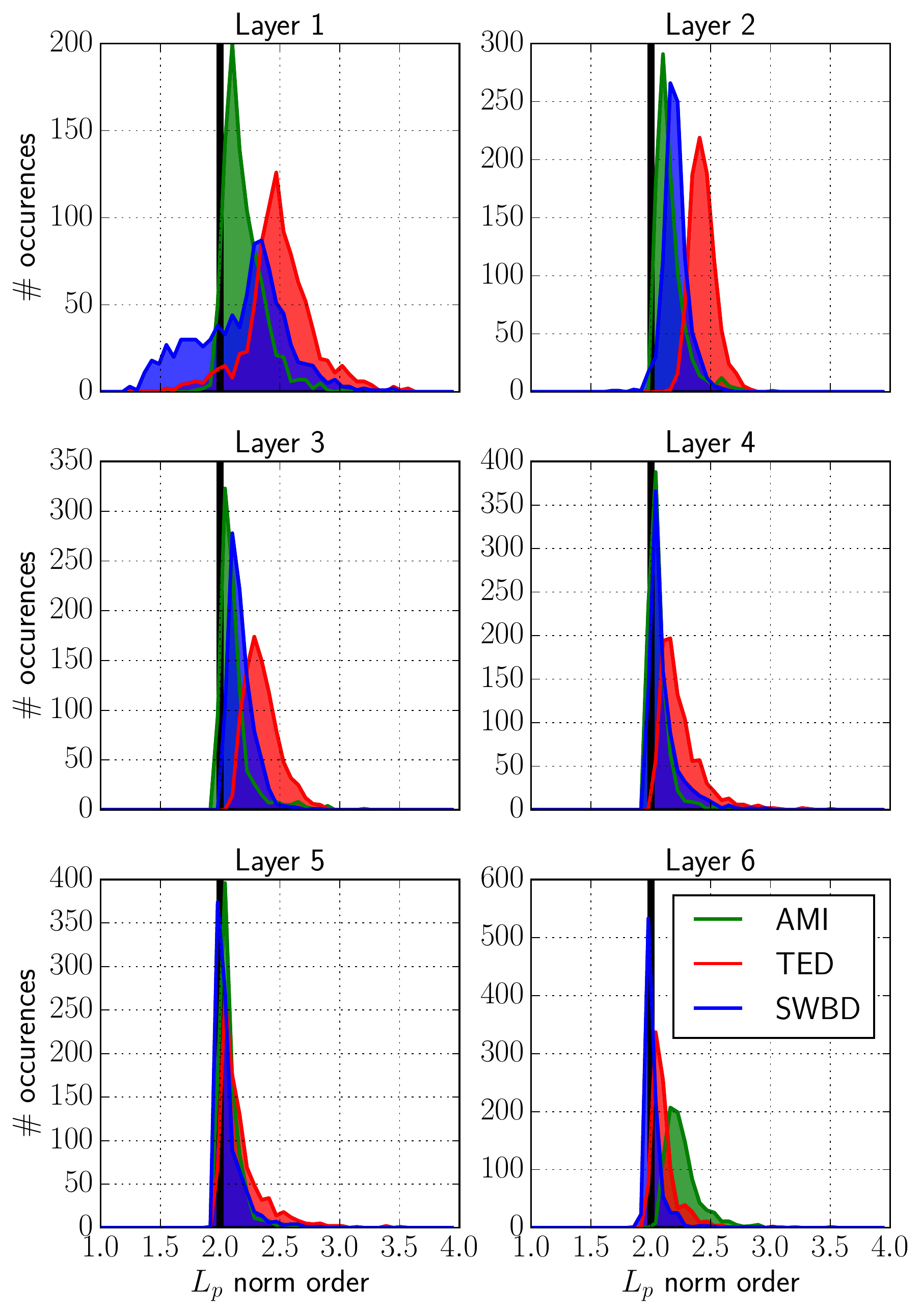}
\caption{$L_p$ orders for the three corpora used in this work. Particular models share the same structure of hidden layers (the same number of $L_p$ units per layer -- 900), though both dimensionality of the output layers as well as the acoustic features used to train each model, are different. Vertical black line at 2 denotes an initial $p$ setting of $L_p$ units.}
\label{fig:lp_orders_data}
\end{figure}

Fig.\ \ref{fig:lp_orders_data} gives more insight into the \diffp models by showing how the final distributions of the learned order $p$ differ across AMI, TED and SWBD corpora. $p$ deviates more from its initialisation in the lower layers of the model; there is also a difference across corpora.  This follows the intuition of how a multi-layer network builds its representation: lower layers are more dependent on acoustic variabilities, normalising for such effects, and hence feature extractors may differ across datasets -- in contrast to the upper layers which rely on features abstracted away from the acoustic data.  For these corpora, the order $p$ rarely exceeded 3, sometimes dropping below 2 -- especially for layer 1 with SWBD data.  However, most $L_p$ units, especially in higher layers, tend to have $p \sim 2$. This corresponds to previous work~\cite{Zhang2014} in which fixed $L_{p=2}$ units tended to obtain lower WER. A similar analysis of \diffg pooling does not show large data-dependent differences in the learned pooling parameters.

\textbf{Training speed:} Table \ref{tab:times} shows the average training speeds for each of the considered models. Training pooling units is significantly more expensive than training baseline DNN models. This is to be expected as the pooling operations cannot be easily and fully vectorised. In our implementation training the \diffg or \diffp models is about 40\% slower than training a baseline DNN. Not optimising $p$ during training \eqref{eq:lp_p} decreases the gap to about 20\% slower. This indicates that training using fixed $L_2$ units, and then adapting the order $p$ in a speaker adaptive manner could make a good compromise.

\begin{table}[t]
\small
\caption{Average training speeds [frames/second] as obtained for each model type on SWBD data and GTX980 GPGPU boards.}
\label{tab:times}
\centerline{
\begin{tabular}{c|c|c|c}
DNN & \gauss & \difff & \diffp \\ 
\hline \hline
9k  &  5.2k  & 7.1k & 5.4k \\ 
 \hline
\end{tabular}}
\end{table}

\subsection{Adaptation experiments} \label{ssec:adapt}
\begin{figure}[b]
\subfigure[]{
  \includegraphics[width=0.465\columnwidth]{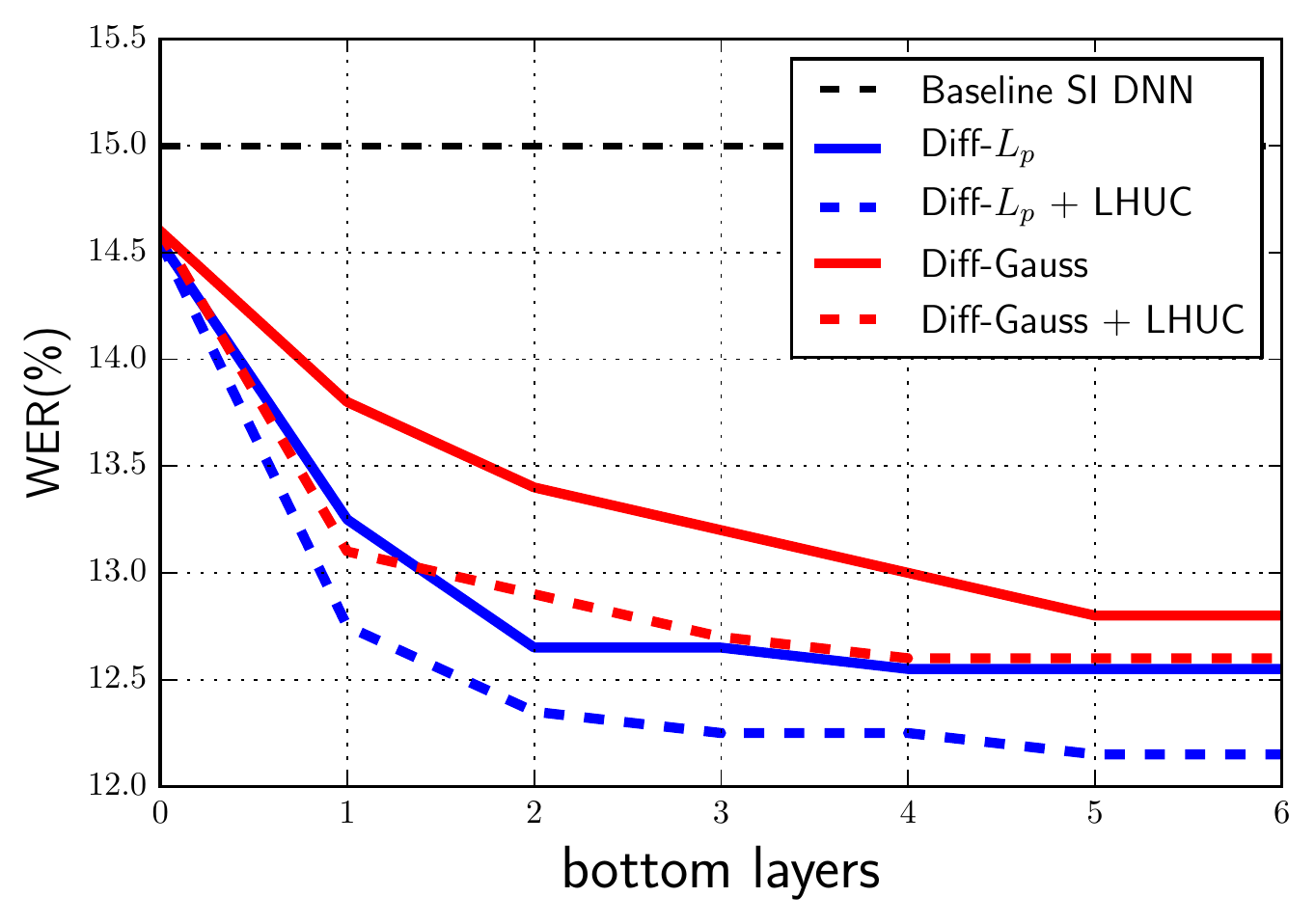}
}
\subfigure[]{
  \includegraphics[width=0.465\columnwidth]{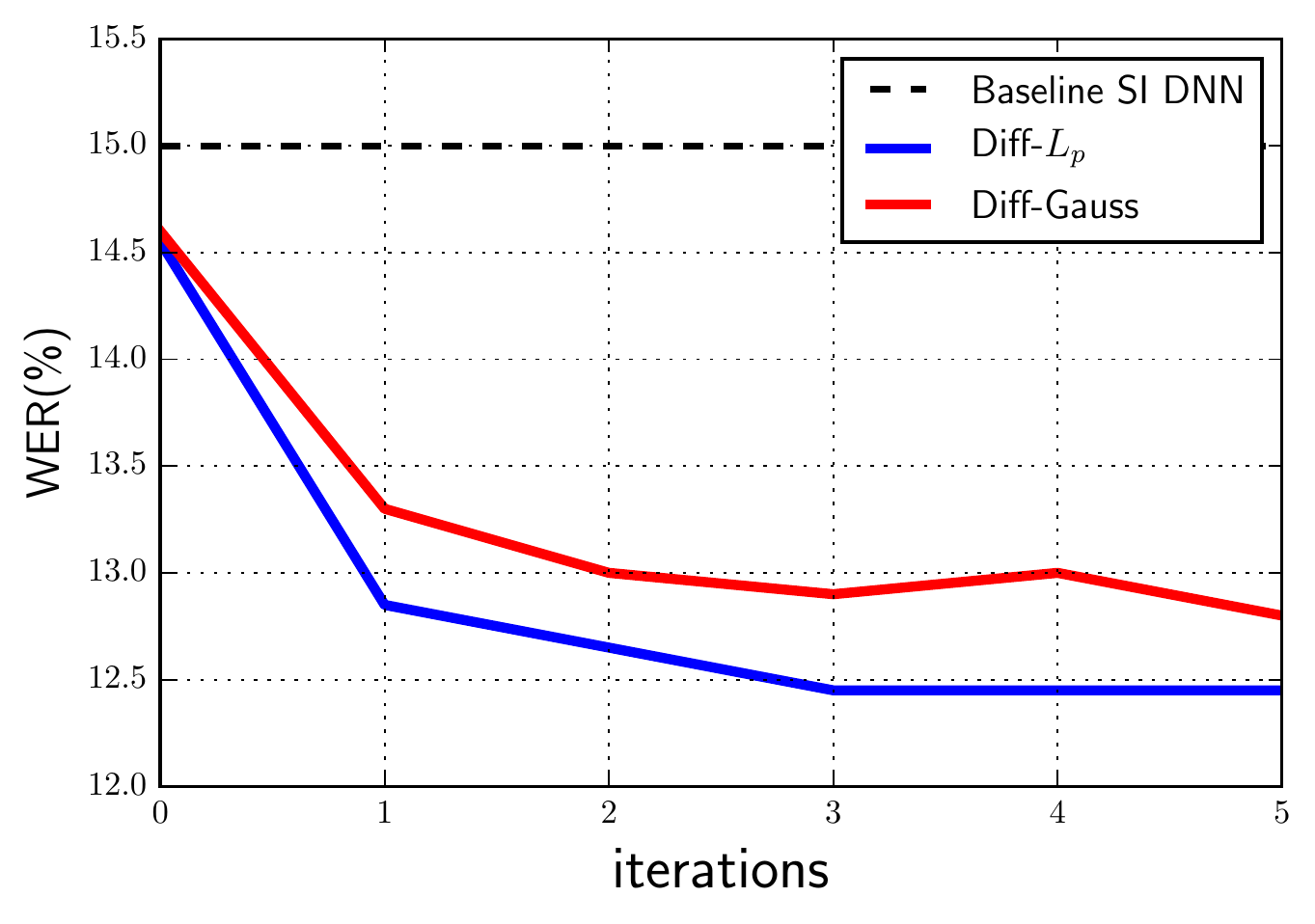}
} 
\vspace{-4mm}
\caption{WER(\%) on \tsta as a function of a) number of bottom layers adapted with pooling operators and (optional) \LHUC transforms and b) number of adaptation iterations}
\label{fig:adapt_stats}
\end{figure}

We initially used the TED talks corpus to investigate how WERs are affected by adapting different layers in the model.  The results indicated that adapting only the bottom layer brings the largest drop in WER; however, adapting more layers further improves the accuracy for both \diffp and \diffg models (Fig.\ \ref{fig:adapt_stats} (a)). Since obtaining the gradients for the pooling parameters at each layer is inexpensive compared to the overall back-propagation, and adapting bottom layer gives largest gains, in the remainder of this work we adapt all pooling units. Similar trends hold when pooling adaptation is combined with \lhuc adaptation, which on \tsta improves the accuracies by 0.2-0.3\% absolute. 

\begin{table}[t]
\small
\caption{WER(\%) results for different subsets of adapted parameters of \diffp model on TED (\tsta), AMI(\amieval) and SWBD (\swbdeval) test sets}
\vspace{-1mm}
\label{tab:lp_params}
\centerline{
\begin{tabular}{l|c||c|c|c}
Model		& \#SD Parameters & TED & AMI  & SWBD  \\ 
\hline \hline 
\diffp		 & -                       &  14.5 &  27.6  & 21.3 \\
+ \lhuc    & \footnotesize{$P(L-1)$} &  12.8 &  25.8  & 20.5 \\ \hline
+\updatep    & \footnotesize{$P(L-1)$} &  12.5 &  25.8  & 20.1 \\ 
++\updatec   &  \footnotesize{$(P+PK)(L-1)$} &  12.3 &  25.5  & 20.5 \\ 
++ \lhuc 	 &  \footnotesize{$2P(L-1)$} &  12.3 &  25.6  & 20.0 \\ 
\hline 
\multicolumn{5}{l}{\footnotesize $L$ - \#layers, $P$ - \#pooling units in layer, $K$ - pool size}
\end{tabular}}
\end{table}

\begin{table}[t]
\small
\caption{WER(\%) results for different subsets of adapted parameters of \diffg model on TED (\tsta), AMI(\amieval) and SWBD (\swbdeval) test-sets.}
\vspace{-1mm}
\label{tab:gauss_params}
\centerline{
\begin{tabular}{l|c||c|c|c}
Model		& \#SD Parameters & TED & AMI  & SWBD  \\ 
\hline \hline 
\gauss 		& - 						& 14.6	&  29.0 & 21.4 \\  
+ \lhuc  	& \footnotesize{$P(L-1)$} 	& 12.8 	&  - 	& - \\ \hline
+\updateu   & \footnotesize{$P(L-1)$} 	& 13.1 	&  - 	& - \\
+\updateb  	& \footnotesize{$P(L-1)$} 	& 13.1 	&  - 	& - \\ 
+\updatee	& \footnotesize{$P(L-1)$} 	& 12.7 	&  - 	& - \\ 
+\updateg  	& \footnotesize{$2P(L-1)$} 	& 12.8	&  27.3 & 20.7 \\ 
++ \lhuc 	& \footnotesize{$3P(L-1)$} 	& 12.5 	&  27.0 & 20.4 \\
++\updatee 	& \footnotesize{$3P(L-1)$} 	& 12.3	&  26.9 & 20.3 \\
\hline 
\multicolumn{5}{l}{\footnotesize $L$ - \#layers, $P$ - \#pooling units in layer}
\end{tabular}}
\vspace{-2mm}
\end{table}

Fig.\ \ref{fig:adapt_stats} (b) shows WER vs. the number of adaptation iterations. The results indicate that one adaptation iteration is sufficient and, more importantly, the model does not overfit when more iterations are used. This suggests that it is not necessary to regularise the model carefully (by Kullback-Leibler divergence~\cite{Yu2013}, for instance) which is usually required when weights that directly transform the data are adapted.  In the remainder, we adapt all models with a learning rate of $0.8$ for three iterations (optimised on \dev).

Table \ref{tab:lp_params} shows the effect of adapting different pooling parameters (including \lhuc amplitudes) for $L_p$ units. Updating only $p$, rather than any other stand-alone pooling parameter, gives a lower WER than \lhuc adaptation with the same number of parameters (cf Fig.\ \ref{fig:diffp_regions}); however, updating both brings further reductions in WER.  Adapting the bias is more data-dependent with a substantial increase in WER for SWBD; this also significantly increases the number of adapted parameters. Hence we adapted either $p$ alone, or $p$ with \lhuc in the remaining experiments

Table \ref{tab:gauss_params} shows similar analysis but for \diffg model.  For \diffg, it is beneficial to update both $\bm \mu$ and $\bm \beta$ (as in~\cite{Swietojanski:ICASSP15}), and \lhuc was also found to be complementary. Notice, adapting with \lhuc scalers is similar to altering $\eta$ in eq.~\eqref{eq:nonlin} (assuming $\eta$ is tied per pool, as mentioned in Section~\ref{sec:gauss}). As such, new parameters need not be introduced to adapt \diffg with \lhuc as it is the case for \diffp units. In fact, last two rows of Table~\ref{tab:gauss_params} show that jointly updating $\mu$, $\beta$ and $\eta$ gives lower WER than updating $\mu$, $\beta$ and applying  $\lhuc$ after pooling (see Fig.~\ref{fig:diffp_nnets}).

\begin{figure}[t]
 \centering  \includegraphics[width=1.0\columnwidth]{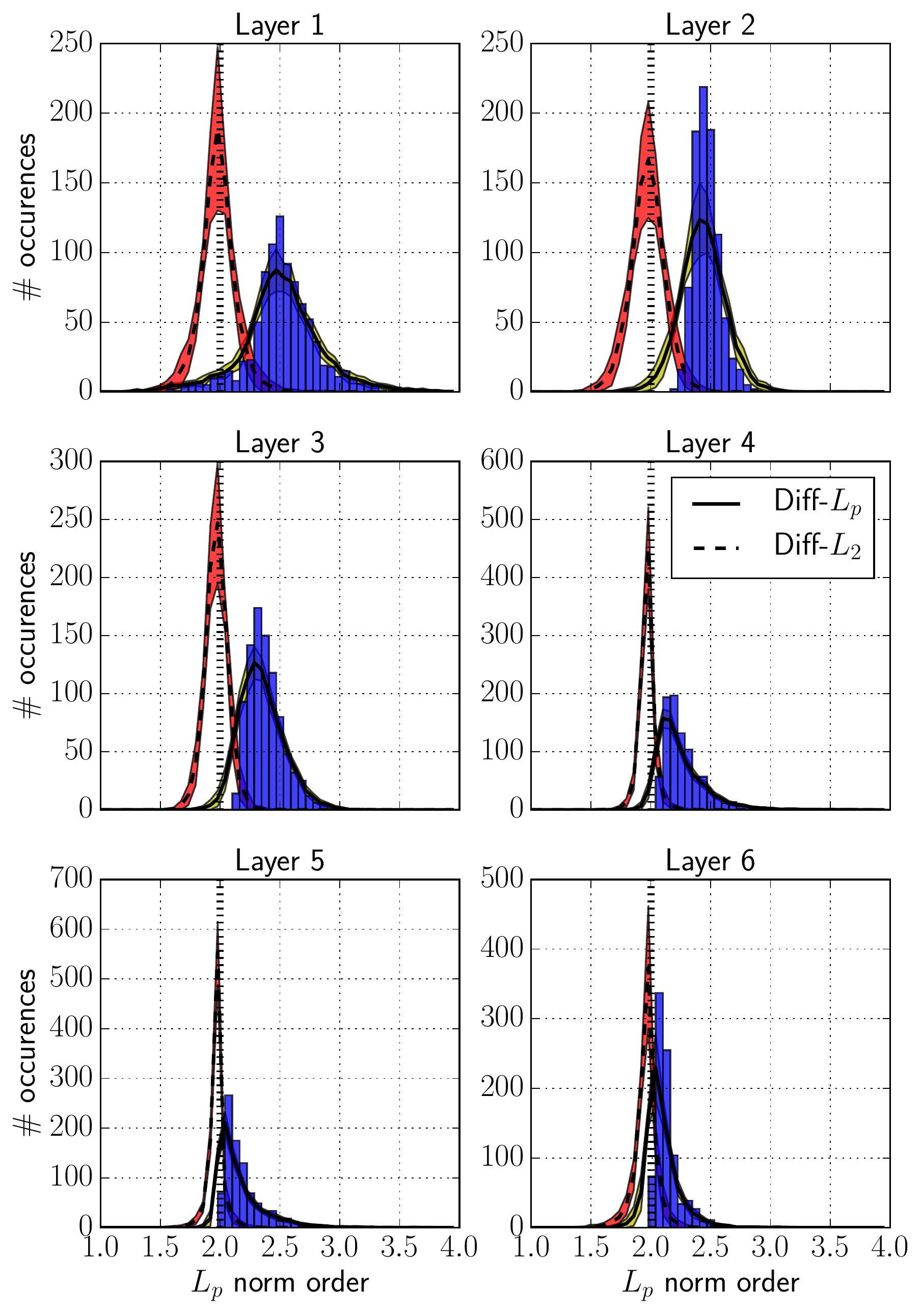}
\caption{Layer-wise histograms of learned $L_p$ orders (speaker independent) on TED. The vertical line (dashed-black) at 2 is the initial value of $p$; the black solid line denotes the mean contour ($\pm$ standard deviations in yellow) of the distribution of $p$ obtained after adaptation to 28 speakers of \tstc. Likewise, the dashed black line is the mean of the adapted $L_p$ orders ($\pm$ standard deviations in red) starting from a fixed-order \difff speaker independent model. (Best viewed in colour.)}
\label{fig:lp_orders}
\end{figure}

\textbf{Analysis of \diffp}: Fig.\ \ref{fig:lp_orders} shows how the distribution of  $p$ changes after the \diffp model adapts to each of the 28 speakers of \tstc. We plot the speaker independent histograms as well as the contours of the mean bin frequencies for each layer. For the adapted models the distributions of $p$ become less dispersed, especially in higher layers, which can be interpreted  as shrinking the decision regions of particular $L_p$ units (cf Fig.\ \ref{fig:diffp_regions}). This follows the intuition that speaker adaptation involves reducing the variability that needs to be modelled, in contrast to the speaker independent model.

Taking into account the increased training time of \diffp models, one can also consider training fixed order \difff~\cite{Zhang2014}, adapting $p$ using \eqref{eq:lp_p}. The results in Fig.\ \ref{fig:lp_orders}, as well as later results, cover this scenario.  The adapted \difff models display a similar trend  in the distribution of $p$ to the \diffp models.

\begin{figure}[t]
 \centering \includegraphics[width=1.0\columnwidth]{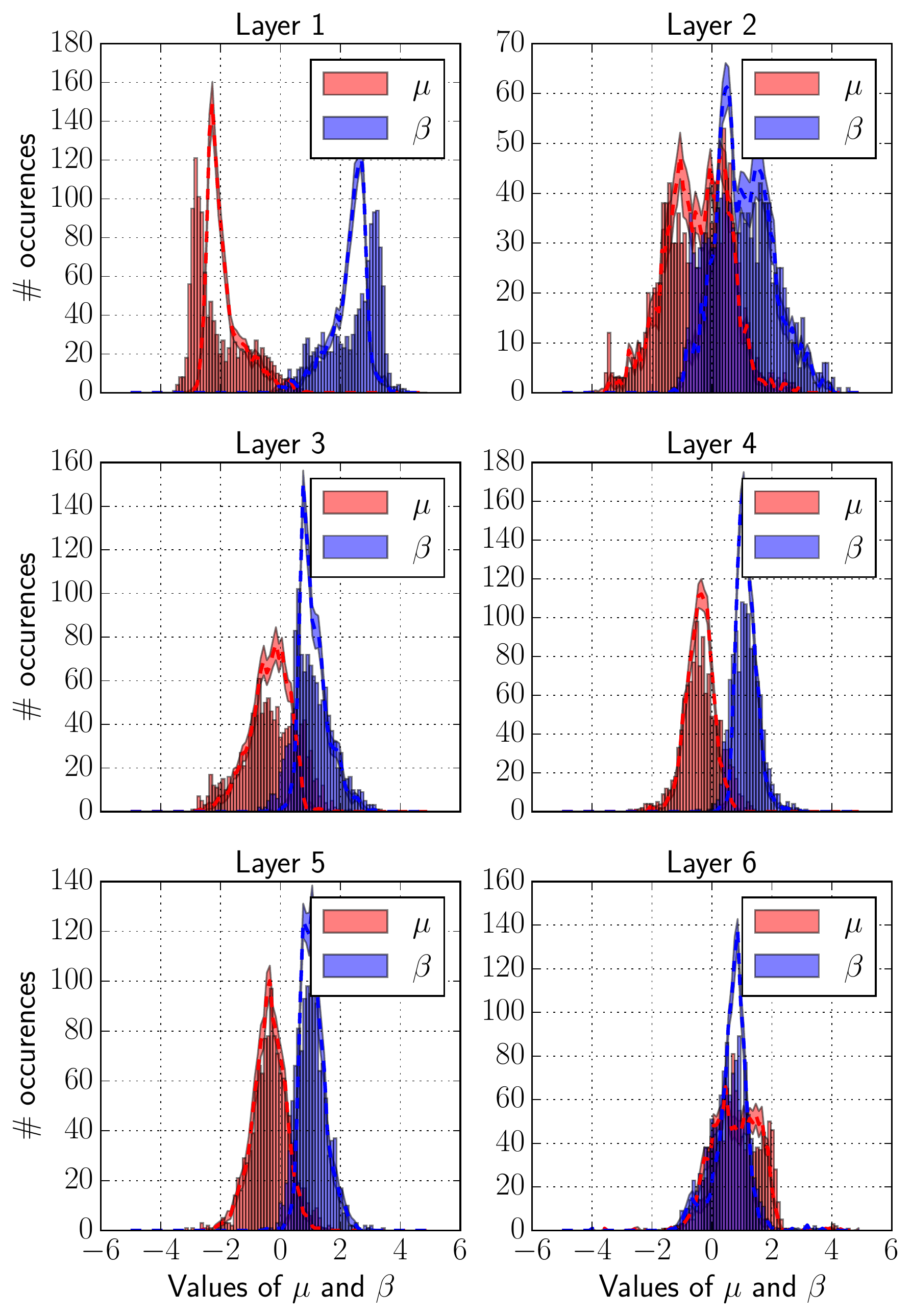}
\caption{Layer-wise histograms of learned \diffg pooling parameters $\{\mu, \beta\}$ during speaker independent training on TED. We also plot the altered mean contours ($\pm$ standard deviation) of the adapted pooling parameters on 28 speakers of \tstc. (Best viewed in color.)}
\label{fig:gauss_orders}
\end{figure}

\textbf{Analysis of \diffg}: We performed a similar investigation on the learned \diffg pooling parameters (Fig.\ \ref{fig:gauss_orders}). In the bottom layers they are characterised by a large negative means and positive precisions which has the effect of turning off many units.
After adaptation, some of them become more active, which can be seen based on shifted distributions of adapted pooling parameters in Fig.\ \ref{fig:gauss_orders}. The adaptation with \diffg has a similar effect as the adaptation of slopes and amplitudes \cite{Zhao2015_slopes, Zhang2015}, but adapts $K$ times fewer parameters.

\begin{figure*}[!htbp]
\center
\subfigure[]{
  \includegraphics[width=0.7\columnwidth]{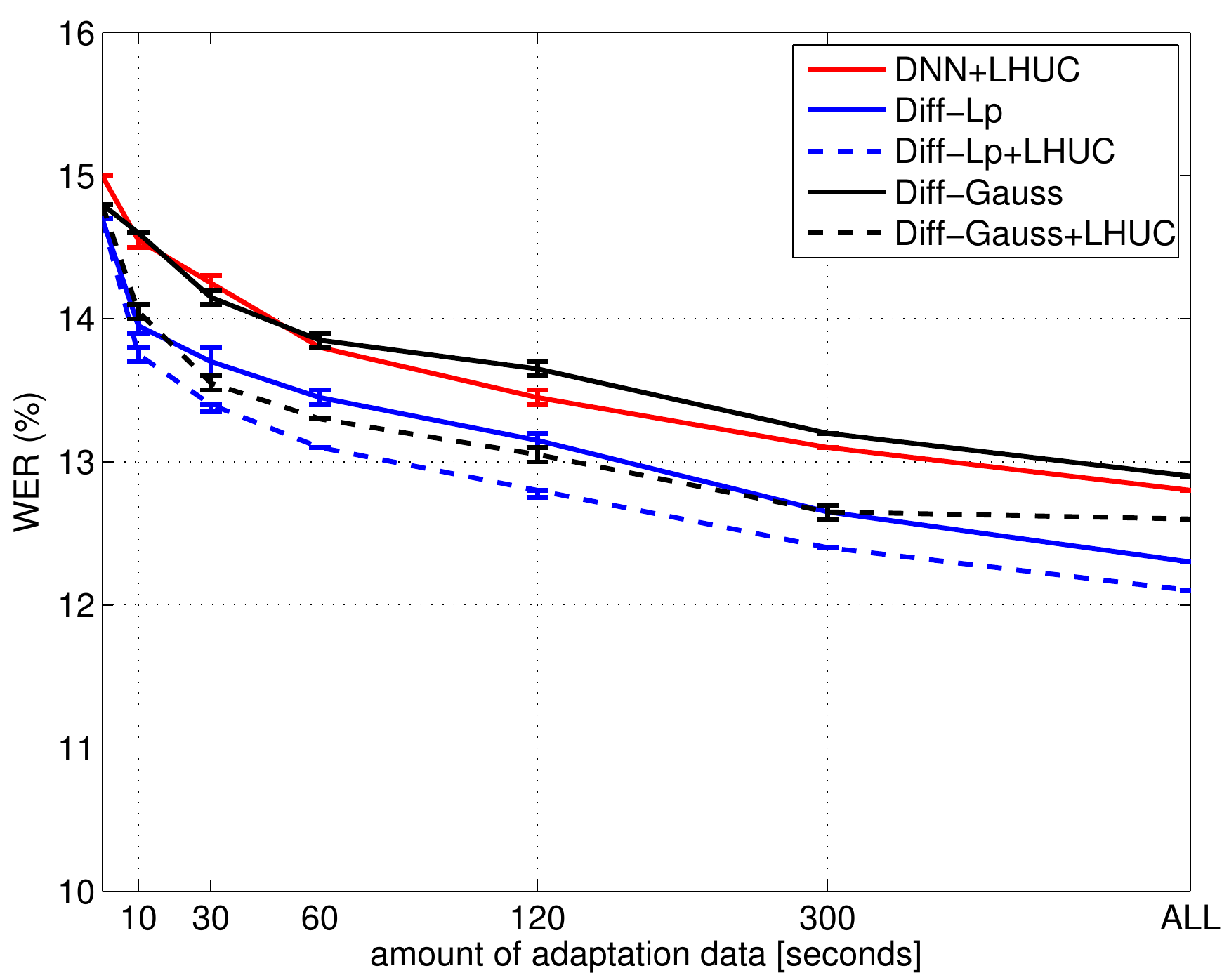}
}
\subfigure[]{
 \includegraphics[width=0.7\columnwidth]{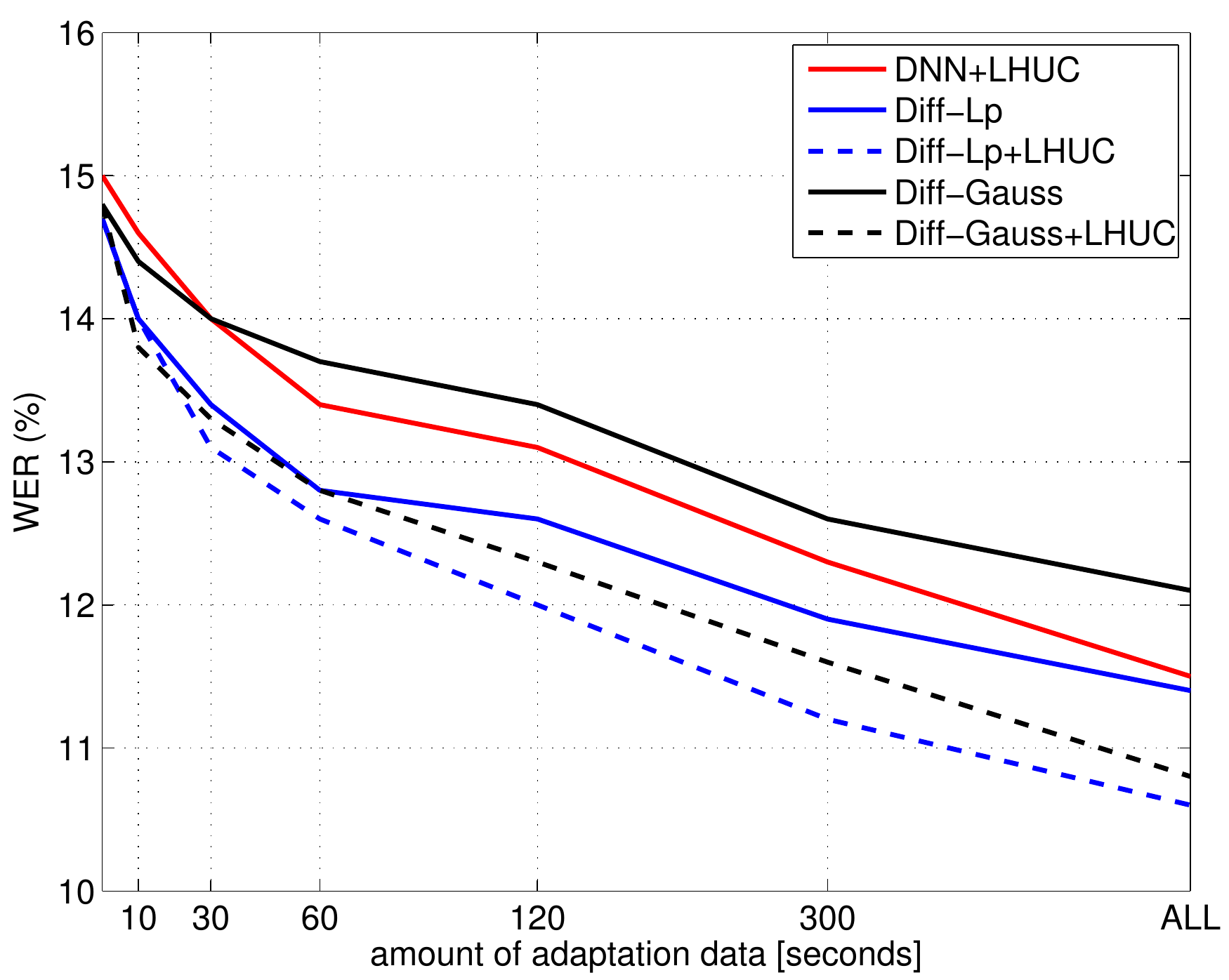}
}
\caption{WER(\%) on \tsta for different amounts of adaptation data with (a) unsupervised and (b) oracle adaptation targets. 
}
\label{fig:as_data}
\end{figure*}

\textbf{Amount of adaptation data and quality of targets}: We investigated the effect of the amount of adaptation data by randomly selecting adaptation utterances from \tsta to give totals of 10s, 30s, 60s, 120s, 300s and more  speaker-specific adaptation data per talker (Fig.\ \ref{fig:as_data} (a)). The WERs are an average over three independent runs, each sampling a different set of adaptation utterances (we did more passes in our previous work~\cite{Swietojanski2014_lhuc, Swietojanski:ICASSP15}, however, both \lhuc and differentiable pooling operators were not sensitive to this aspect, resulting in small error bars between different results obtained with different random utterances). The \diffp models offer lower WER and more rapid adaptation, with 10s of adaptation data resulting in a decrease in WER by 0.6\% absolute (3.6\% relative) which further increases up to 2.1\% absolute (14.4\% relative) when using all the speaker's data in an unsupervised manner. \diffg is comparable in terms of WER to a DNN adapted with \lhuc. In addition, both methods are complementary to \lhuc adaptation, and to feature-space adaptation with fMLLR (Tables \ref{tab:tedwers} and \ref{tab:swbdwers}).

In order to demonstrate the modelling capacities of the different model-based adaptation techniques, we carried out a supervised adaptation (oracle) experiment in which the adaptation targets were obtained by aligning the audio data with reference transcripts (Fig.\ \ref{fig:as_data} (b)).  We do not refine what the model knows about speech, nor the way it classifies it (the feature receptors and output layer are fixed during adaptation and remain speaker independent), but show that the re-composition and interpolation of these basis functions to approximate the unseen distribution of adaptation data is able to decrease the WER by 26.7\% relative for \diffp + \LHUC scenario.

The methods are also not very sensitive to the quality of adaptation targets, they show very similar trends as \lhuc, for which exact results for different qualities of adaptation targets resulting from re-scoring adaptation hypotheses with different language models were reported in~\cite{swietojanski2016lhuc}.



\textbf{Summary}: Results for the proposed techniques are summarised in Tables \ref{tab:amiwers}, \ref{tab:tedwers}, and \ref{tab:swbdwers} for AMI, TED, and SWBD, respectively. The overall observed trends are as follows: (I) speaker independent pooling models return lower WERs than the baseline DNNs: \diffg $<$ \difff $\leq$ \diffp (although the last two seem to be data-dependent); (II) the pooling models (\diffg, \difff and \diffp) are complementary to both fMLLR and \lhuc adaptation -- as expected, the final gain depends on the degree of data mismatch; (III) one can effectively train speaker independent \difff models and later alter $p$ in a speaker dependent manner; (IV) the average relative improvement across all tasks with respect to baseline unadapted DNN models were 6.8\% for \diffg, 9.1\% for \difff and 10.4\% for \diffp; and (V) when comparing \lhuc adapted DNN to \lhuc adapted differentiable pooling models, the relative reductions in WER for the pooling models were  2\%, 3.4\% and 4.8\%  for \diffg, \difff and \diffp, respectively.


\begin{table}[t]
\small
\caption{WER(\%) on AMI - Individual Headset Microphones and AM trained on FBANK features}
\label{tab:amiwers}
\centerline{
\begin{tabular}{l||c|c}
Model &  \amidev & \amieval \\ 
\hline \hline
DNN 	&   26.8 		& 29.1   \\ 
+\lhuc 	&  	25.6 	& 27.1   \\ \hline 
\gauss 	&  	26.7 		& 29.0   \\ 
+\updateg &  26.0 	& 27.3   \\ 
++\lhuc 	&  25.7 	& 27.0   \\  \hline
\flpf   &  26.1 & 28.5  \\
+\updatep 	&  25.5		& 26.9   \\
++\lhuc 	&  25.3		&  26.7  \\ \hline
\llpf 	&  	 25.4 	& 27.6   \\ 
+\updatep 	& 24.7 	& 25.8   \\
++\lhuc 	& 24.7 	& 25.6   \\
\hline
\end{tabular}}
\end{table}

\begin{table}[t]
\small
\caption{Summary WER(\%) results on TED test sets from IWSLT12 and IWSLT13 evaluations.}
\label{tab:tedwers}
\centerline{
\begin{tabular}{l||c|c|c|c}
Model & \footnotesize{\dev} & \footnotesize{\tsta} & \footnotesize{\tstb} & \footnotesize{\tstc} \\ 
\hline \hline
DNN			& 15.4 & 15.0  & 12.1 & 22.1  \\ 
+\lhuc      & 14.5 & 12.8  & 11.0  & 19.2 \\ 
+fMLLR      & 14.5 & 12.9  & 10.9  & 20.8 \\ 
++\lhuc     & 14.1 & 11.8 & 10.3    & 18.4 \\ 
\hline \hline
\gauss 		&  15.4 & 14.6	 & 11.9  &  21.8  \\ 
+\updateg   &  14.5  & 12.8  &  11.2 & 19.5  \\ 
++\lhuc  	&  14.1  & 12.5  	& 10.8  &  18.7  \\ \hline
+fMLLR      &  14.6	& 13.1	&  10.9  & 21.1  \\ 
++\updateg     &  14.3	& 12.4	&  10.7 & 19.4  \\ 
+++\lhuc  	&  14.1	& 12.1 &  10.5  & 18.9  \\ 
 \hline \hline
\difff		& 15.0 &  14.6 & 11.8 & 21.7  \\ 
+\updatep   & 14.1 &  12.6 & 11.0 & 18.5 \\ 
++\lhuc    	& 13.9 &  12.3 & 10.8 & 18.1 \\ \hline
\diffp		& 14.9 &  14.5 & 11.7 & 21.6  \\ 
+\updatep   & 14.2 &  12.5 & 10.8 & 18.4 \\ 
++\lhuc    	& 14.0 &  12.2 & 10.6 & 17.9 \\ \hline
+fMLLR      & 14.0 &  12.5 & 10.6 & 20.3 \\
++\updatep  & 13.7 &  11.5 & 10.0 & 18.0  \\
+++\lhuc    & 13.4 &  11.4 & 9.8  & 17.6 \\
 \hline
\end{tabular}}
\end{table}

\begin{table}[t]
\footnotesize
\caption{Summary WER(\%) results on Switchboard \swbdeval} \vspace{-1mm}
\label{tab:swbdwers}
\centerline{
\begin{tabular}{l||c|c|c}
& \multicolumn{3}{c}{\swbdeval} \\ \cline{2-4}
Model & SWB & CHE & TOTAL \\ 
\hline \hline
\multicolumn{4}{c}{Baseline models} \\ \hline
DNN  	&  	15.8 & 28.4 		& 22.1   \\ 
+\lhuc 	&  	15.4 & 27.0 		& 21.2   \\
+\fmllr	&  	14.3 & 26.1 		& 20.3   \\ 
++\lhuc &  	14.2 & 25.6 		& 19.9   \\ \hline
\multicolumn{4}{c}{\diffg models} \\ \hline
\gauss 	& 15.1 & 27.8	 & 21.4		\\
+\updateg 	& 14.8 & 26.6	 & 20.7		\\  
++\lhuc 	& 14.6 & 26.2	 & 20.4	  \\ 
+\fmllr	& 14.4 & 26.1	 & 20.3		\\ 
++\updateg 	& 14.3 & 25.5	 & 19.9		\\ \hline
\multicolumn{4}{c}{\difff models} \\ \hline
\difff  	&  	14.9 & 28.0 		& 21.3   \\ 
+\updatep  	&  	14.2 & 26.0 		& 20.1   \\ 
++\lhuc  	& 14.2 & 25.9	 & 20.1   \\ 
+\fmllr 	&  	13.9 & 25.5 		& 19.7   \\ 
++\updatep &  	13.5 & 24.9 		& 19.2   \\ \hline
\multicolumn{4}{c}{\diffp models} \\ \hline
\diffp  	&  	14.8 & 28.0 		& 21.3   \\
+\updatep 	&  	14.2 & 26.0 		& 20.1   \\
++\lhuc  	& 14.1 & 25.9	 & 20.0   \\
+\fmllr  	&  	13.7 & 25.3 		& 19.5   \\
++\updatep  	&  	13.5 & 24.6 		& 19.0   \\ \hline
\end{tabular}}
\vspace{-2mm}
\end{table}

\section{Discussion and Conclusions} \label{sec:conc}

We have proposed the use of differentiable pooling operators with DNN acoustic models to perform unsupervised speaker adaptation. Differentiable pooling operators offer a relatively-low dimensional set of parameters which may be adapted in a speaker-dependent fashion.

We investigated the complementarity of differentiable pooling adaptation with two other approaches -- model-based \lhuc adaptation and feature-space fMLLR adaptation. We have not performed an explicit comparison with an i-vector approach to adaptation. However, some recent papers have compared i-vector adaptation with either \lhuc and/or fMLLR on similar data which enables us some make  indirect comparisons. For example, Samarakoon and Sim~\cite{Samarakoon:ICASSP16} showed that speaker-adaptive training with i-vectors gives a comparable results to test-only \lhuc using TED data, and Miao et. al~\cite{Miao2015} suggested that \lhuc is better than a standard use of i-vectors (as in Saon et al.~\cite{Saon2013}) on TED data, with a more sophisticated i-vector post-processing needed to equal \lhuc. Since the proposed \diffp and \diffg techniques resulted in WERs that were at least as good as \lhuc (and were found to be complementary to fMLLR) we conclude that the proposed pooling-based adaptation techniques are competitive. 

In the future, one could investigate  extending the proposed techniques to speaker adaptive training (SAT)~\cite{Anastasakos1996, gales2000cluster}, for example in a similar spirit as proposed in the context of SAT-LHUC~\cite{Swietojanski:ICASSP16}. In addition it would be interesting to investigate the suitability of adapting pooling regions in the framework of sequence discriminative training~\cite{povey2005discriminative,Kingsbury2009,Vesely:IS13}. Our experience of \lhuc in this framework~\cite{swietojanski2016lhuc}, together with the observation that the pooling models are not prone to over-fitting in the case of small amounts of adaptation data, suggests that adaptation based on differentiable pooling is a promising technique for sequence trained models.

\section*{Acknowledgement}

The NST research data collection may be accessed at http://datashare.is.ed.ac.uk/handle/10283/786. This research utilised a K40 GPGPU board donated by NVIDA Corporation. The authors would like to thank the reviewers for insightful comments that helped to improve the manuscript.

\ifCLASSOPTIONcaptionsoff
  \newpage
\fi

\bibliographystyle{IEEEbib}
\bibliography{master}

\begin{thebibliography}{10}

\bibitem{Hinton2012}
G~Hinton, L~Deng, D~Yu, GE~Dahl, A~Mohamed, N~Jaitly, A~Senior, V~Vanhoucke,
  P~Nguyen, TN~Sainath, and B~Kingsbury,
\newblock ``Deep neural networks for acoustic modeling in speech recognition:
  The shared views of four research groups,''
\newblock {\em IEEE Signal Processing Magazine}, vol. 29, no. 6, pp. 82--97,
  Nov 2012.

\bibitem{Yu2013_repr}
D~Yu, M~Seltzer, J~Li, J-T Huang, and F~Seide,
\newblock ``Feature learning in deep neural networks - studies on speech
  recognition,''
\newblock in {\em Proc. ICLR}, 2013.

\bibitem{Neto1995}
J~Neto, L~Almeida, M~Hochberg, C~Martins, L~Nunes, S~Renals, and T~Robinson,
\newblock ``Speaker adaptation for hybrid {HMM--ANN} continuous speech
  recognition system,''
\newblock in {\em Proc. Eurospeech}, 1995, pp. 2171--2174.

\bibitem{li2010lin}
B~Li and KC~Sim,
\newblock ``Comparison of discriminative input and output transformations for
  speaker adaptation in the hybrid {NN/HMM} systems,''
\newblock in {\em Proc. Interspeech}, 2010.

\bibitem{jan2010speaker}
J~Trmal, J~Zelinka, and L~M{\"u}ller,
\newblock ``On speaker adaptive training of artificial neural networks,''
\newblock in {\em Proc. Interspeech}, 2010.

\bibitem{Seide2011}
F~Seide, X~Chen, and D~Yu,
\newblock ``Feature engineering in context-dependent deep neural networks for
  conversational speech transcription,''
\newblock in {\em Proc. IEEE ASRU}, 2011.

\bibitem{Yao2012}
K~Yao, D~Yu, F~Seide, H~Su, L~Deng, and Y~Gong,
\newblock ``Adaptation of context-dependent deep neural networks for automatic
  speech recognition.,''
\newblock in {\em Proc. IEEE SLT}, 2012, pp. 366--369.

\bibitem{Swietojanski2013}
P~Swietojanski, A~Ghoshal, and S~Renals,
\newblock ``Revisiting hybrid and {GMM-HMM} system combination techniques,''
\newblock in {\em Proc. IEEE ICASSP}, 2013, pp. 6744--6748.

\bibitem{Yu2013}
D~Yu, K~Yao, H~Su, G~Li, and F~Seide,
\newblock ``{KL}-divergence regularized deep neural network adaptation for
  improved large vocabulary speech recognition.,''
\newblock in {\em Proc. IEEE ICASSP}, 2013, pp. 7893--7897.

\bibitem{Abdel-Hamid2013_is}
O~{Abdel-Hamid} and H~Jiang,
\newblock ``Rapid and effective speaker adaptation of convolutional neural
  network based models for speech recognition.,''
\newblock in {\em Proc. ICSA Interspeech}, 2013, pp. 1248--1252.

\bibitem{Swietojanski2014_lhuc}
P~Swietojanski and S~Renals,
\newblock ``Learning hidden unit contributions for unsupervised speaker
  adaptation of neural network acoustic models,''
\newblock in {\em Proc. IEEE SLT}, 2014.

\bibitem{Gales1998}
MJF Gales,
\newblock ``Maximum likelihood linear transformations for {HMM}-based speech
  recognition,''
\newblock {\em Computer Speech and Language}, vol. 12, pp. 75--98, April 1998.

\bibitem{young1994tied}
SJ~Young and PC~Woodland,
\newblock ``{S}tate clustering in hidden {M}arkov model-based continuous speech
  recognition,''
\newblock {\em Computer Speech and Language}, vol. 8, no. 4, pp. 369--383,
  1994.

\bibitem{Bourlard1994}
H~Bourlard and N~Morgan,
\newblock {\em Connectionist Speech Recognition: A Hybrid Approach},
\newblock Kluwer Academic Publishers, 1994.

\bibitem{Renals1994}
S~Renals, N~Morgan, H~Bourlard, M~Cohen, and H~Franco,
\newblock ``Connectionist probability estimators in {HMM} speech recognition,''
\newblock {\em IEEE Transactions on Speech and Audio Processing}, vol. 2, pp.
  161--174, 1994.

\bibitem{Bridle1990softmax}
JS~Bridle,
\newblock ``Probabilistic interpretation of feedforward classification network
  outputs, with relationships to statistical pattern recognition,''
\newblock in {\em Neurocomputing}, F~{Fogelman Souli\'e} and J~H\'erault, Eds.,
  pp. 227--236. Springer, 1990.

\bibitem{Nair2010}
V~Nair and G~Hinton,
\newblock ``Rectified linear units improve restricted {B}oltzmann machines,''
\newblock in {\em Proc. ICML}, 2010, pp. 131--136.

\bibitem{Hermansky2000}
H~Hermansky, DPW Ellis, and S~Sharma,
\newblock ``Tandem connectionist feature extraction for conventional {HMM}
  systems,''
\newblock in {\em Proc. IEEE ICASSP}, 2000, pp. 1635--1638.

\bibitem{Grezl2007}
F~Grezl, M~Karafiat, S~Kontar, and J~Cernocky,
\newblock ``Probabilistic and bottle-neck features for {LVCSR} of meetings,''
\newblock in {\em Proc. IEEE ICASSP}, 2007, pp. IV--757--IV--760.

\bibitem{Mohamed2011}
A~Mohamed, TN~Sainath, G~Dahl, B~Ramabhadran, GE~Hinton, and MA~Picheny,
\newblock ``Deep belief networks using discriminative features for phone
  recognition,''
\newblock in {\em Proc. IEEE ICASSP}, 2011, pp. 5060--5063.

\bibitem{Hain2012}
T~Hain, L~Burget, J~Dines, PN~Garner, F~Gr\'{e}zl, A~{El Hannani},
  M~Karaf\'iat, M~Lincoln, and V~Wan,
\newblock ``Transcribing meetings with the {AMIDA} systems,''
\newblock {\em IEEE Transactions on Audio, Speech and Language Processing},
  vol. 20, pp. 486--498, 2012.

\bibitem{Sainath2012}
TN~Sainath, B~Kingsbury, and B~Ramabhadran,
\newblock ``Auto-encoder bottleneck features using deep belief networks.,''
\newblock in {\em Proc. IEEE ICASSP}, 2012, pp. 4153--4156.

\bibitem{Sainath2013_cnns}
TN~Sainath, A~Mohamed, B~Kingsbury, and B~Ramabhadran,
\newblock ``Deep convolutional neural networks for lvcsr,''
\newblock in {\em Proc. IEEE ICASSP}, 2013, pp. 8614--8618.

\bibitem{Bell2013_mlan}
P~Bell, P~Swietojanski, and S~Renals,
\newblock ``Multi-level adaptive networks in tandem and hybrid {ASR} systems,''
\newblock in {\em Proc. IEEE ICASSP}, 2013.

\bibitem{yoshioka2014investigation}
T~Yoshioka, A~Ragni, and MJF Gales,
\newblock ``Investigation of unsupervised adaptation of dnn acoustic models
  with filter bank input,''
\newblock in {\em Proc. IEEE ICASSP}, 2014, pp. 6344--6348.

\bibitem{Abrash1995}
V~Abrash, H~Franco, A~Sankar, and M~Cohen,
\newblock ``Connectionist speaker normalization and adaptation,''
\newblock in {\em Proc. Eurospeech}, 1995, pp. 2183--2186.

\bibitem{Dehak2010}
N~Dehak, PJ~Kenny, R~Dehak, P~Dumouchel, and P~Ouellet,
\newblock ``Front end factor analysis for speaker verification,''
\newblock {\em IEEE Trans Audio, Speech and Language Processing}, vol. 19, pp.
  788--798, 2010.

\bibitem{Karafiat2011}
M~Karafiat, L~Burget, P~Matejka, O~Glembek, and J~Cernozky,
\newblock ``{I-vector}-based discriminative adaptation for automatic speech
  recognition,''
\newblock in {\em Proc. IEEE ASRU}, 2011.

\bibitem{Saon2013}
G~Saon, H~Soltau, D~Nahamoo, and M~Picheny,
\newblock ``Speaker adaptation of neural network acoustic models using
  i-vectors.,''
\newblock in {\em Proc. IEEE ASRU}, 2013, pp. 55--59.

\bibitem{senior2014adapt}
A~Senior and I~Lopez-Moreno,
\newblock ``Improving {DNN} speaker independence with i-vector inputs.,''
\newblock in {\em Proc. IEEE ICASSP}, 2014, pp. 225--229.

\bibitem{Gupta2014}
V~Gupta, P~Kenny, P~Ouellet, and T~Stafylakis,
\newblock ``{I}-vector based speaker adaptation of deep neural networks for
  french broadcast audio transcription,''
\newblock in {\em Proc. IEEE ICASSP}, 2014.

\bibitem{Karanasou2014}
P~Karanasou, Y~Wang, MJF Gales, and PC~Woodland,
\newblock ``Adaptation of deep neural network acoustic models using factorised
  i-vectors,''
\newblock in {\em Proc. ICSA Interspeech}, 2014, pp. 2180--2184.

\bibitem{Miao2015}
Y~Miao, H~Zhang, and F~Metze,
\newblock ``Speaker adaptive training of deep neural network acoustic models
  using i-vectors,''
\newblock {\em IEEE/ACM Transactions on Audio, Speech, and Language
  Processing}, vol. 23, no. 11, pp. 1938--1949, Nov 2015.

\bibitem{Samarakoon:ICASSP16}
L~Samarakoon and K~C Sim,
\newblock ``On combining i-vectors and discriminative adaptation methods for
  unsupervised speaker normalization in dnn acoustic models,''
\newblock in {\em Proc. IEEE ICASSP}, 2016, pp. 5275--5279.

\bibitem{Liu2014}
Y~Liu, P~Zhang, and T~Hain,
\newblock ``Using neural network front-ends on far field multiple microphones
  based speech recognition,''
\newblock in {\em Proc. IEEE ICASSP}, 2014, pp. 5542--5546.

\bibitem{Bridle1990}
JS~Bridle and S~Cox,
\newblock ``Recnorm: Simultaneous normalisation and classification applied to
  speech recognition,''
\newblock in {\em Advances in Neural Information and Processing Systems}, 1990.

\bibitem{Abdel-Hamid2013}
O~Abdel-Hamid and H~Jiang,
\newblock ``Fast speaker adaptation of hybrid {NN/HMM} model for speech
  recognition based on discriminative learning of speaker code,''
\newblock in {\em Proc. IEEE ICASSP}, 2013, pp. 4277--4280.

\bibitem{Xue2014_scodes}
S~Xue, O~Abdel-Hamid, J~Hui, L~Dai, and Q~Liu,
\newblock ``Fast adaptation of deep neural network based on discriminant codes
  for speech recognition,''
\newblock {\em IEEE/ACM Transactions on Audio, Speech, and Language
  Processing}, vol. 22, no. 12, pp. 1713--1725, Dec 2014.

\bibitem{kundu2016factor}
S~Kundu, G~Mantena, Y~Qian, T~Tan, M~Delcroix, and KC~Sim,
\newblock ``Joint acoustic factor learning for robust deep neural network based
  automatic speech recognition,''
\newblock in {\em Proc. IEEE ICASSP}, March 2016, pp. 5025--5029.

\bibitem{Liao2013}
H~Liao,
\newblock ``Speaker adaptation of context dependent deep neural networks.,''
\newblock in {\em Proc. IEEE ICASSP}, 2013, pp. 7947--7951.

\bibitem{Huang2015}
Y~Huang and Y~Gong,
\newblock ``Regularized sequence-level deep neural network model adaptation,''
\newblock in {\em Proc. ICSA Interspeech}, 2015, pp. 1081--1085.

\bibitem{Ochiai2014}
T~Ochiai, S~Matsuda, X~Lu, C~Hori, and S~Katagiri,
\newblock ``Speaker adaptive training using deep neural networks,''
\newblock in {\em Proc. IEEE ICASSP}, 2014, pp. 6349--6353.

\bibitem{Sini2013}
SM~Siniscalchi, J~Li, and CH~Lee,
\newblock ``Hermitian polynomial for speaker adaptation of connectionist speech
  recognition systems,''
\newblock {\em IEEE Trans Audio, Speech, and Language Processing}, vol. 21, pp.
  2152--2161, 2013.

\bibitem{Zhao2015_slopes}
Y~Zhao, J~Li, J~Xue, and Y~Gong,
\newblock ``Investigating online low-footprint speaker adaptation using
  generalized linear regression and click-through data,''
\newblock in {\em Proc. IEEE ICASSP}, 2015, pp. 4310--4314.

\bibitem{Swietojanski:ICASSP16}
P~Swietojanski and S~Renals,
\newblock ``{SAT-LHUC}: Speaker adaptive training for learning hidden unit
  contributions,''
\newblock in {\em Proc. IEEE ICASSP}, 2016, pp. 5010--5014.

\bibitem{huang2015maximum}
Z~Huang, S~M Siniscalchi, I-F Chen, J~Wu, and C-H Lee,
\newblock ``Maximum a-posteriori adaptation of network parameters in deep
  models,''
\newblock {\em arXiv preprint arXiv:1503.02108}, 2015.

\bibitem{Huang2015_mt}
Z~Huang, J~Li, SM~Siniscalchi, I-F Chen, J~Wu, and C-H Lee,
\newblock ``Rapid adaptation for deep neural networks through multi-task
  learning,''
\newblock in {\em Proc. ICSA Interspeech}, 2015, pp. 3625--3629.

\bibitem{Swietojanski2015_mt}
P~Swietojanski, P~Bell, and S~Renals,
\newblock ``{Structured output layer with auxiliary targets for
  context-dependent acoustic modelling},''
\newblock in {\em Proc. ICSA Interspeech}, 2015, pp. 3605--3609.

\bibitem{Price2014}
R~Price, K~Iso, and K~Shinoda,
\newblock ``Speaker adaptation of deep neural networks using a hierarchy of
  output layers,''
\newblock in {\em Proc. IEEE SLT}, 2014.

\bibitem{Swietojanski:ICASSP15}
P~Swietojanski and S~Renals,
\newblock ``Differentiable pooling for unsupervised speaker adaptation,''
\newblock in {\em Proc. IEEE ICASSP}, 2015, pp. 4305--4309.

\bibitem{hubel1962receptive}
D~Hubel and T~Wiesel,
\newblock ``Receptive fields, binocular interaction, and functional
  architecture in the cat's visual cortex,''
\newblock {\em Journal of Physiology}, vol. 160, pp. 106--154, 1962.

\bibitem{Fukushima1982}
K~Fukushima and S~Miyake,
\newblock ``Neocognitron: A new algoriothm for pattern recognition tolerant of
  deformations,''
\newblock {\em Pattern Recognition}, vol. 15, pp. 455--469, 1982.

\bibitem{LeCun1989}
Y~{LeCun}, B~Boser, JS~Denker, D~Henderson, RE~Howard, W~Hubbard, and
  LD~Jackel,
\newblock ``Backpropagation applied to handwritten zip code recognition,''
\newblock {\em Neural Computation}, vol. 1, pp. 541--551, 1989.

\bibitem{LeCun1998a}
Y~{LeCun}, L~Bottou, Y~Bengio, and P~Haffner,
\newblock ``Gradient-based learning applied to document recognition,''
\newblock {\em Proceedings of the IEEE}, vol. 86, pp. 2278--2324, 1998.

\bibitem{Riesenhuber1999}
M~Riesenhuber and T~Poggio,
\newblock ``Hierarchical models of object recognition in cortex,''
\newblock {\em Nature Neuroscience}, vol. 2, pp. 1019--1025, 1999.

\bibitem{Ranzato2007}
MA~Ranzato, FJ~Huang, Y-L Boureau, and Y~{LeCun},
\newblock ``Unsupervised learning of invariant feature hierarchies with
  applications to object recognition,''
\newblock in {\em Proc. IEEE CVPR}, 2007.

\bibitem{Boureau2010}
Y-L Boureau, J~Ponce, and Y~{LeCun},
\newblock ``A theoretical analysis of feature pooling in visual recognition,''
\newblock in {\em Proc. ICML}, 2010.

\bibitem{Goodfellow2013}
IJ~Goodfellow, D~Warde-Farley, M~Mirza, A~Courville, and Y~Bengio,
\newblock ``Maxout networks,''
\newblock in {\em Proc. ICML}, 2013, pp. 1319--1327.

\bibitem{Miao2013b}
Y~Miao, F~Metze, and S~Rawat,
\newblock ``Deep maxout networks for low-resource speech recognition,''
\newblock in {\em Proc. IEEE ASRU}, 2013.

\bibitem{Cai2013}
M~Cai, Y~Shi, and J~Liu,
\newblock ``Deep maxout neural networks for speech recognition,''
\newblock in {\em Proc. IEEE ASRU}, 2013, pp. 291--296.

\bibitem{swietojanski2014maxout}
P~Swietojanski, J~Li, and J-T Huang,
\newblock ``Investigation of maxout networks for speech recognition,''
\newblock in {\em Proc. IEEE ICASSP}, 2014.

\bibitem{Renals2014}
S~Renals and P~Swietojanski,
\newblock ``Neural networks for distant speech recognition,''
\newblock in {\em Proc. HSCMA}, 2014.

\bibitem{Toth2014}
L~Toth,
\newblock ``Convolutional deep maxout networks for phone recognition,''
\newblock in {\em Proc. ICSA Interspeech}, 2014.

\bibitem{Zeiler2012}
MD~Zeiler and R~Fergus,
\newblock ``Differentiable pooling for hierarchical feature learning,''
\newblock {\em CoRR}, vol. abs/1207.0151, 2012.

\bibitem{Sermanet2012}
P~Sermanet, S~Chintala, and Y~{LeCun},
\newblock ``Convolutional neural networks applied to house numbers digit
  classification,''
\newblock {\em CoRR}, vol. abs/1204.3968, 2012.

\bibitem{Sainath2013}
TN~Sainath, B~Kingsbury, A~Mohamed, GE~Dahl, G~Saon, H~Soltau, T~Beran,
  AY~Aravkin, and B~Ramabhadran,
\newblock ``Improvements to deep convolutional neural networks for {LVCSR},''
\newblock in {\em Proc. IEEE ASRU}, 2013, pp. 315--320.

\bibitem{Zhang2014}
X~Zhang, J~Trmal, D~Povey, and S~Khudanpur,
\newblock ``Improving deep neural network acoustic models using generalized
  maxout networks,''
\newblock in {\em Proc. IEEE ICASSP}, 2014.

\bibitem{swietojanski2016lhuc}
P~Swietojanski, J~Li, and S~Renals,
\newblock ``Learning hidden unit contributions for unsupervised acoustic model
  adaptation,''
\newblock {\em IEEE/ACM Transactions on Audio, Speech, and Language
  Processing}, vol. 24, no. 8, pp. 1450--1463, 2016.

\bibitem{Caglar2013}
C~G{\"{u}}l{\c{c}}ehre, K~Cho, R~Pascanu, and Y~Bengio,
\newblock ``Learned-norm pooling for deep feedforward and recurrent neural
  networks,''
\newblock in {\em Proc. ECML and KDD}. 2014, pp. 530--546, Springer-Verlag.

\bibitem{rumelhart1986learning}
DE~Rumelhart, GE~Hinton, and RJ~Williams,
\newblock ``Learning internal representations by error-propagation,''
\newblock in {\em Parallel Distributed Processing}, vol.~1, pp. 318--362. MIT
  Press, 1986.

\bibitem{Cettolo2012}
M~Cettolo, C~Girardi, and M~Federico,
\newblock ``Wit$^3$: Web inventory of transcribed and translated talks,''
\newblock in {\em Proc. EAMT}, 2012, pp. 261--268.

\bibitem{godfrey1992switchboard}
JJ~Godfrey, EC~Holliman, and J~McDaniel,
\newblock ``{SWITCHBOARD}: Telephone speech corpus for research and
  development,''
\newblock in {\em Proc. IEEE ICASSP}. IEEE, 1992, pp. 517--520.

\bibitem{Carletta_LRE2007}
J~Carletta,
\newblock ``Unleashing the killer corpus: Experiences in creating the
  multi-everything {AMI} meeting corpus.,''
\newblock {\em Language Resources and Evaluation}, vol. 41, no. 2, pp.
  181--190, 2007.

\bibitem{Renals_ASRU2007}
S~Renals, T~Hain, and H~Bourlard,
\newblock ``Recognition and understanding of meetings: The {AMI} and {AMIDA}
  projects,''
\newblock in {\em Proc. IEEE ASRU}, Kyoto, 12 2007,
\newblock IDIAP-RR 07-46.

\bibitem{Dahl2012}
GE~Dahl, D~Yu, L~Deng, and A~Acero,
\newblock ``Context-dependent pre-trained deep neural networks for
  large-vocabulary speech recognition,''
\newblock {\em IEEE Transaction on Audio, Speech, and Language Processing},
  vol. 20, no. 1, pp. 30--42, 2012.

\bibitem{Bell2014}
P~Bell, P~Swietojanski, J~Driesen, M~Sinclair, F~McInnes, and S~Renals,
\newblock ``The {UEDIN} system for the {IWSLT} 2014 evaluation,''
\newblock in {\em Proc. IWSLT}, 2014, pp. 26--33.

\bibitem{Swietojanski_ASRU13}
P~Swietojanski, A~Ghoshal, and S~Renals,
\newblock ``Hybrid acoustic models for distant and multichannel large
  vocabulary speech recognition,''
\newblock in {\em Proc. IEEE ASRU}, 2013.

\bibitem{cieri2004}
C~Cieri and D~Millerand~K Walker,
\newblock ``The {Fisher} corpus: a resource for the next generations of
  speech-to-text,''
\newblock in {\em Proc LREC}, 2004.

\bibitem{Vesely:IS13}
K~Vesely, A~Ghoshal, L~Burget, and D~Povey,
\newblock ``Sequence-discriminative training of deep neural networks,''
\newblock in {\em Proc. ICSA Interspeech}, 2013, pp. 2345--2349.

\bibitem{Kaldi:ASRU11}
D~Povey, A~Ghoshal, G~Boulianne, L~Burget, O~Glembek, N~Goel, M~Hannemann,
  P~Motl\'{i}\v{c}ek, Y~Qian, P~Schwarz, J~Silovsk\'{y}, G~Stemmer, and
  K~Vesel\'{y},
\newblock ``The {Kaldi} speech recognition toolkit,''
\newblock in {\em Proc. IEEE ASRU}, December 2011.

\bibitem{Renals1992}
S~Renals, N~Morgan, M~Cohen, and H~Franco,
\newblock ``Connectionist probability estimation in the {DECIPHER} speech
  recognition system,''
\newblock in {\em Proc. IEEE ICASSP}, 1992.

\bibitem{srivastava2014dropout}
N~Srivastava, G~Hinton, A~Krizhevsky, I~Sutskever, and R~Salakhutdinov,
\newblock ``Dropout: A simple way to prevent neural networks from
  overfitting,''
\newblock {\em Journal of Machine Learning Research}, vol. 15, pp. 1929--1958,
  2014.

\bibitem{Zhang2015}
C~Zhang and PC~Woodland,
\newblock ``Parameterised sigmoid and {ReLU} hidden activation functions for
  {DNN} acoustic modelling,''
\newblock in {\em Proc. ICSA Interspeech}, 2015, pp. 3224--3228.

\bibitem{Anastasakos1996}
T~Anastasakos, J~McDonough, R~Schwartz, and J~Makhoul,
\newblock ``A compact model for speaker-adaptive training,''
\newblock in {\em Proc. ICSLP}, 1996, pp. 1137--1140.

\bibitem{gales2000cluster}
MJF Gales,
\newblock ``Cluster adaptive training of hidden markov models,''
\newblock {\em Speech and Audio Processing, IEEE Transactions on}, vol. 8, no.
  4, pp. 417--428, 2000.

\bibitem{povey2005discriminative}
D~Povey,
\newblock {\em Discriminative training for large vocabulary speech
  recognition},
\newblock Ph.D. thesis, University of Cambridge, 2003.

\bibitem{Kingsbury2009}
B~Kingsbury,
\newblock ``Lattice-based optimization of sequence classification criteria for
  neural-network acoustic modeling,''
\newblock in {\em Proc. IEEE ICASSP}, 2009, pp. 3761--3764.

\end{thebibliography}

\begin{IEEEbiography}[{\includegraphics[width=1.1in,height=2.25in,clip,keepaspectratio]{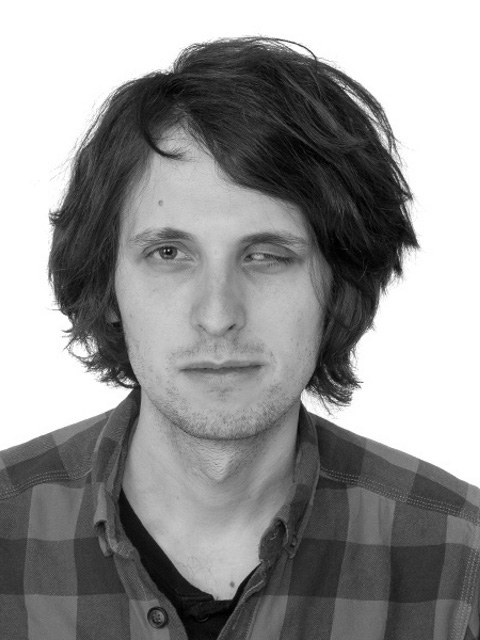}}]{Pawel Swietojanski} received the M.Sc. degree from the AGH University of Science and Technology, Cracow, Poland, and is currently working toward the Ph.D. degree in informatics at the Centre for Speech Technology Research, School of Informatics, University of Edinburgh, U.K. His main research interests include machine learning and its applications to speech processing, with a particular focus on learning representations for acoustic modelling in speech recognition.
\end{IEEEbiography}

\begin{IEEEbiography}[{\includegraphics[width=1.1in,height=2.25in,clip,keepaspectratio]{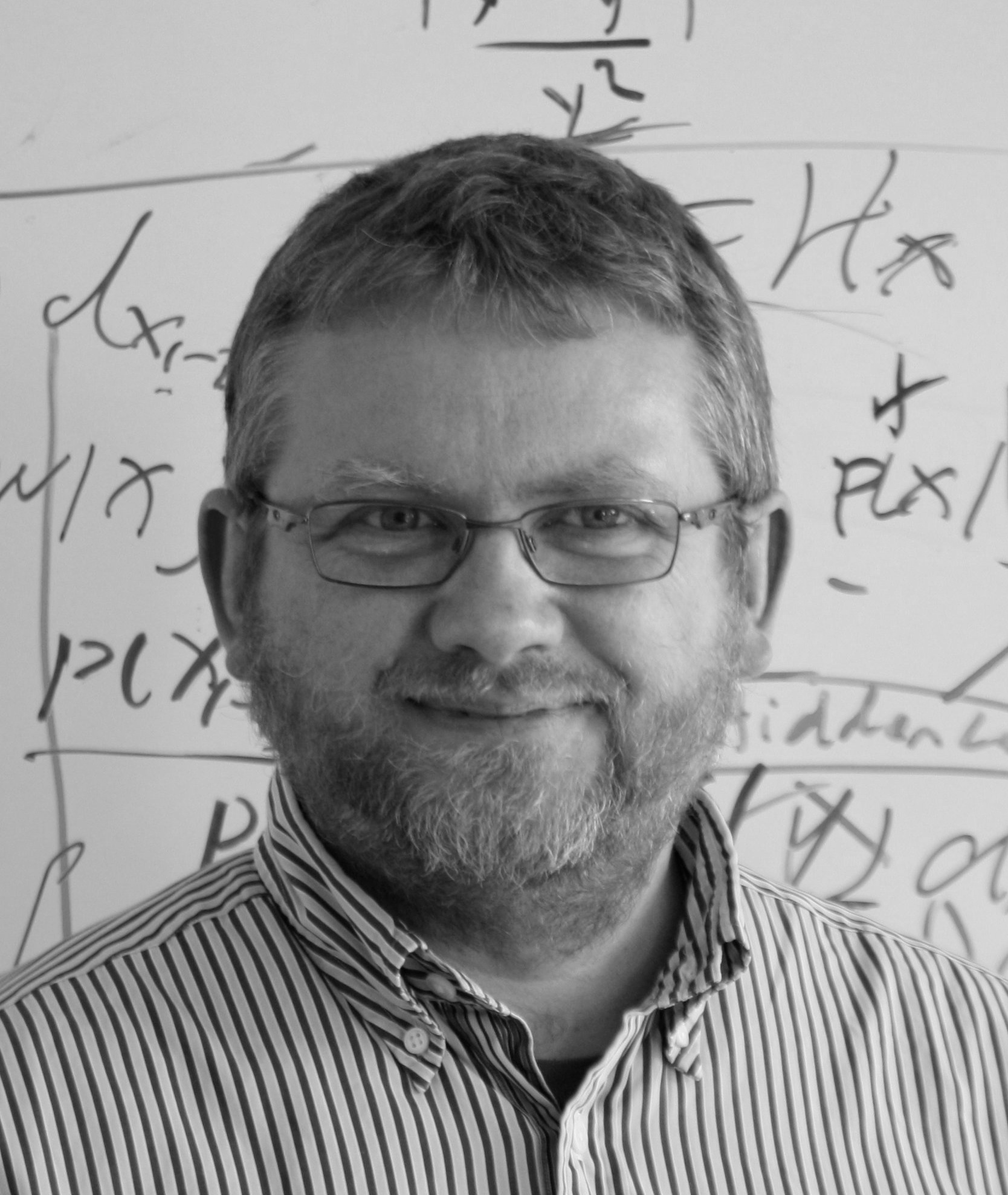}}]{Steve Renals} (M'91 --- SM'11 -- F'14) received the B.Sc. degree from the University of Shefﬁeld, Shefﬁeld, U.K., and the M.Sc. and Ph.D. degrees from Edinburgh. He is a Professor of speech technology at the University of Edinburgh, Edinburgh, U.K. He has previously had positions at ICSI Berkeley, the University of Cambridge, and the University of Shefﬁeld. He is a Senior Area Editor of IEEE/ACM Transactions on Audio, Speech and Language Processing.
\end{IEEEbiography}

\end{document}